\documentclass[10pt,twocolumn,letterpaper]{article}

\usepackage{iccv}
\usepackage{times}
\usepackage{epsfig}
\usepackage{graphicx}
\usepackage{amsmath}
\usepackage{amssymb}
\usepackage{subfigure}
\usepackage{multirow}
\usepackage{booktabs}
\usepackage{color}
\usepackage[pagebackref=true,breaklinks=true,letterpaper=true,colorlinks,bookmarks=false]{hyperref}

\iccvfinalcopy % *** Uncomment this line for the final submission

\def\iccvPaperID{6349} % *** Enter the ICCV Paper ID here

\ificcvfinal\fi

\begin{document}

%%%%%%%%% TITLE
\title{Adaptive Graph Convolution for Point Cloud Analysis}

\author{\renewcommand{\thefootnote}{\fnsymbol{footnote}}Haoran Zhou$^{1}$ \ \ \ Yidan Feng$^{2}$ \ \ \ Mingsheng Fang$^{1}$ \ \ \ Mingqiang Wei$^{2}$\thanks{Co-corresponding authors (mqwei@nuaa.edu.cn/lutong@nju.edu.cn)} \\
	Jing Qin$^{3}$ \ \ \ Tong Lu$^{1*}$ \\
	$^{1}$Nanjing University \ \ \ $^{2}$Nanjing University of Aeronautics and Astronautics \\
	$^{3}$The Hong Kong Polytechnic University\\
}

\maketitle
\ificcvfinal\thispagestyle{empty}\fi

%%%%%%%%% ABSTRACT
\begin{abstract}
Convolution on 3D point clouds that generalized from 2D grid-like domains is widely researched yet far from perfect. The standard convolution characterises feature correspondences indistinguishably among 3D points, presenting an intrinsic limitation of poor distinctive feature learning. In this paper, we propose Adaptive Graph Convolution (AdaptConv) which generates adaptive kernels for points according to their dynamically learned features. Compared with using a fixed/isotropic kernel, AdaptConv improves the flexibility of point cloud convolutions, effectively and precisely capturing the diverse relations between points from different semantic parts. Unlike popular attentional weight schemes, the proposed AdaptConv implements the adaptiveness inside the convolution operation instead of simply assigning different weights to the neighboring points. Extensive qualitative and quantitative evaluations show that our method outperforms state-of-the-art point cloud classification and segmentation approaches on several benchmark datasets. Our code is available at {\small \url{https://github.com/hrzhou2/AdaptConv-master}}.
\end{abstract}

\section{Introduction}
Point cloud is a standard output of 3D sensors, \eg, LiDAR scanners and RGB-D cameras; it is considered as the simplest yet most efficient shape representation for 3D objects. 
A variety of applications arise with the fast advance of 3D point cloud acquisition techniques, including robotics \cite{rusu2008towards}, autonomous driving \cite{liang2018deep,wang2018deep} and high-level semantic analysis \cite{tchapmi2017segcloud,landrieu2018large}.
%As have been widely used in various applications, including robotics \cite{rusu2008towards}, autonomous driving \cite{liang2018deep,wang2018deep} and high-level semantic analysis \cite{tchapmi2017segcloud,landrieu2018large}, the number of point clouds is growing rapidly---the outdoor/indoor scenes, and even the commodities we consume, are routinely 3D-scanned.
%
Recent years have witnessed considerable attempts to generalize convolutional neural networks (CNNs) to point cloud data for 3D understanding. 
However, unlike 2D images, which are organized as regular grid-like structures, 3D points are unstructured and unordered, discretely distributed on the underlying surface of a sampled object.
%
%This characteristic makes it difficult to harness traditional CNNs for point cloud analysis. 

One common approach is to convert point clouds into regular volumetric representations and hence traditional convolution operations can be naturally applied to them \cite{maturana2015voxnet, riegler2017octnet}. 
Such a scheme, however, often introduces excessive memory cost and is difficult to capture fine-grained geometric details. 
In order to handle the irregularity of point clouds without conversions, PointNet \cite{QiSMG17} applies multi-layer perceptrons (MLPs) independently on each point, which is one of the pioneering works to directly process sparse 3D points.
%Recent years have witnessed a varity of research efforts that are designed specifically to handle the irreguarity of point clouds, directly manipulating raw point cloud data rather than passing to an intermediate regular representation*. This was pioneered by PointNet \cite{QiSMG17} which applies multi-layer perceptron (MLP) independently on each point and subsequently uses a symmetric function to aggregate global feature. These techniques largely treat the points separately using a shared function to maintain permutation invariance which, however, neglects the geometric relationships between points.

\begin{figure}[t]
	\includegraphics[width=0.99\linewidth]{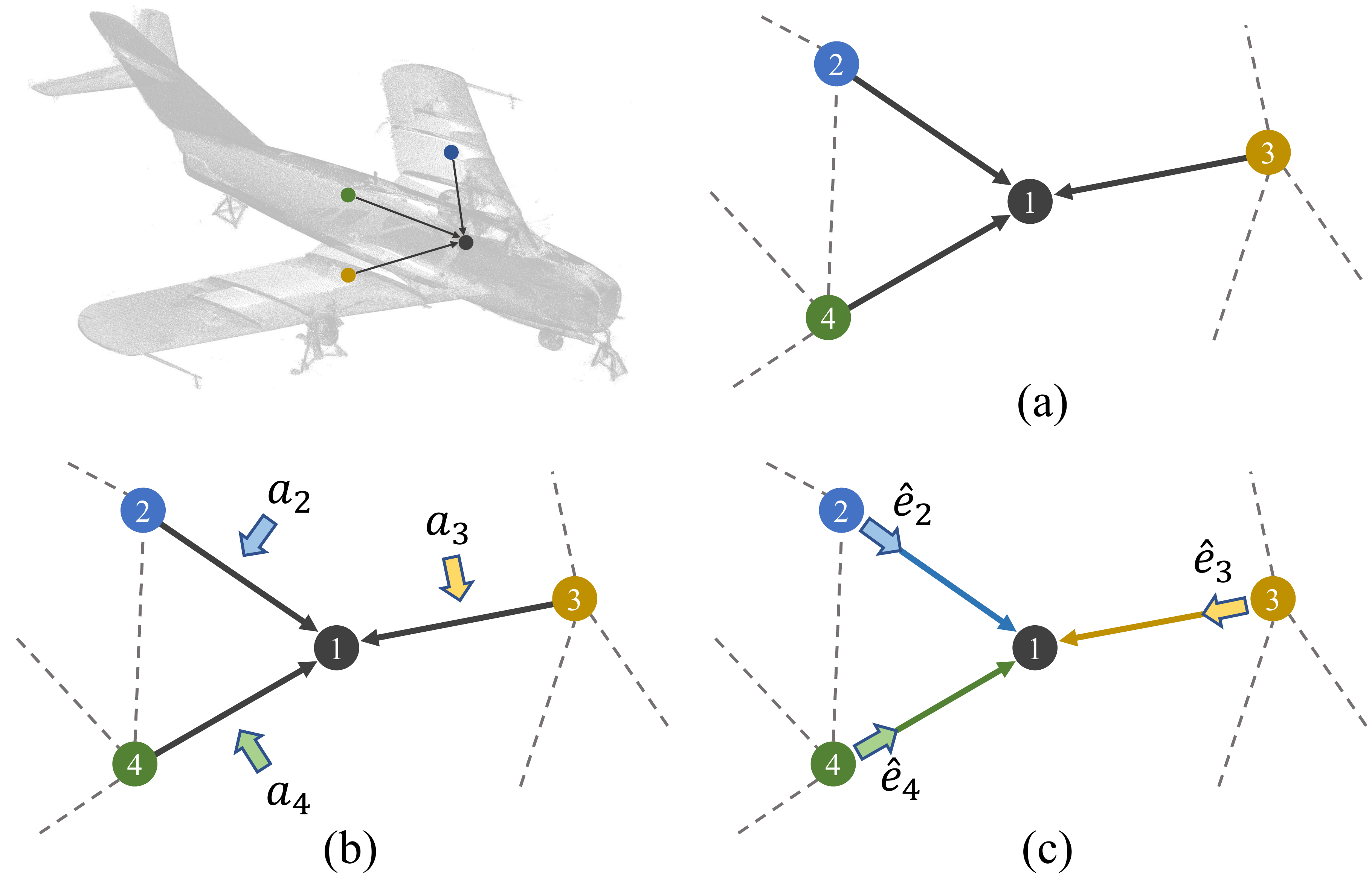}
	\caption{Illustration of adaptive kernels and fixed kernels in the convolution. (a) The standard graph convolution applies a fixed/isotropic kernel (black arrow) to compute features for each point indistinguishably. (b) Based on these features, several attentional weights $a_i$ are assigned to determine their importance. (c) Differently, AdaptConv generates an adaptive kernel $\hat{e}_i$ that is unique to the learned features of each point.}
	\label{fig:intro}
\end{figure}

More recently, several researches have been proposed to utilize the graph-like structures for point cloud analysis. 
Graph CNNs \cite{wang2019dynamic,lin2020convolution,wang2019graph,fujiwara2020neural,hamilton2017inductive} represent a point cloud as graph data according to the spatial/feature similarity between points, and generalize 2D convolutions on images to 3D data. 
In order to process an unordered set of points with varying neighborhood sizes, standard graph convolutions harness shared weight functions over each pair of points to extract the corresponding edge feature. 
This leads to a fixed/isotropic convolution kernel, which is applied identically to all point pairs while neglecting their different feature correspondences. 
Intuitively, for points from different semantic parts of the point cloud (see the neighboring points in Fig.~\ref{fig:intro}), the convolution kernel should be able to distinguish them and determine their different contributions. 

To address this drawback, several approaches \cite{wang2019graph,velivckovic2017graph} are proposed inspired by the idea of attention mechanism \cite{bahdanau2014neural,gehring2016convolutional}. 
As shown in Fig.~\ref{fig:intro} (b), proper attentional weights $a_i$ corresponding to the neighboring points are assigned, trying to identify their different importance when performing the convolution.
%focus on specific part of the local field. 
%
However, these methods are, in principle, still based on the fixed kernel convolution, as the attentional weights are just applied to the features obtained similarly (see the black arrows in Fig.~\ref{fig:intro} (b)). 
%
%In this regard, attentional convolutions cannot solve the inherent limitations of current graph convolutions.%, making it still difficult to capture the delicate geometric features of a point by considering its structural connections to its neighboring points distinctively rather than uniformly. 
%Thus, it is inherently limited for capturing structural connections between points belonging to separate parts of the object.
Considering the intrinsic isotropy of current graph convolutions, these attempts are still limited for detecting the most relevant part in the neighborhood.

Differently, we propose to adaptively establish the relationship between a pair of points according to their learned features. This adaptiveness represents the diversity of kernels unique to each pair of points rather than relying on the predefined weights. 
To achieve this, in this paper, we propose a new graph convolution operator, named \emph{adaptive graph convolution} (AdaptConv). AdaptConv generates adaptive kernels $\hat{e}_i$ for points in the convolution which replace the aforementioned isotropic kernels (see Fig.~\ref{fig:intro} (c)). The key contribution of our work is that the proposed AdaptConv is employed inside the graph convolution rather a weight function that is based on the resulting feature. Furthermore, we explore several choices for the feature convolving design, offering more flexibility to the implementation of the adaptive convolution. Extensive experiments demonstrate the effectiveness of the proposed AdaptConv, achieving state-of-the-art  performances in both classification and segmentation tasks on several benchmark datasets.

\label{sec:introduction}

\section{Related Work}
%Although achieving tremendous success in 2D grid-like structures, deep learning is still not well explored for 3D point cloud analysis. We will review previous researches categorized as voxelization-based, projection-based, graph-based and point-based methods.

\textbf{Voxelization-based and multi-view methods.}
The voxelization/projection strategy has been explored as a simple way in point cloud analysis to build proper representations for adapting the powerful CNNs in 2D vision. A number of works \cite{maturana2015voxnet,wu20153d,le2018pointgrid,wang2017cnn} project point clouds onto regular grids, but inevitably suffer from information loss and enormous computational cost. To alleviate these problems, OctNet \cite{riegler2017octnet} and Kd-Net \cite{klokov2017escape} attempt to use more efficient data structures and skip the computations on empty voxels. Alternatively, the multi-view-based methods \cite{kalogerakis20173d,su2015multi} treat a point cloud as a set of 2D images projected from multiple views, so as to directly leverage 2D CNNs for subsequent processing. However, it is fundamentally difficult to apply these methods to large-scale scanned data, considering the struggle of covering the entire scene from single-point perspectives.

\textbf{Point-based methods.} In order to handle the irregularity of point clouds, state-of-the-art deep networks are designed to directly manipulate raw point cloud data, instead of introducing an intermediate representation. In this way, PointNet \cite{QiSMG17} first proposes to use MLPs independently on each point and subsequently aggregate global features through a symmetric function. Thanks to this design, PointNet is invariant to input point orders, but fails to encode local geometric information, which is important for semantic segmentation tasks. To solve this issue, PointNet++ \cite{qi2017pointnet++} proposes to apply PointNet layers locally in a hierarchical architecture to capture regional information. Alternatively, Huang et al. \cite{huang2018recurrent} sorts unordered 3D points into an ordered list and employs Recurrent Neural Networks (RNN) to extract features according to different dimensions. %In order to process a set of points that are unordered and discrete, there also exists researches that sort the 3D points into an ordered list.\cite{klokov2017escape,gadelha2018multiresolution} proposes to apply Kd-tree to build a 1D list for points according to their coordinates. Although alleviating the unstructured problem, the sorting process is critical to the weight functions, and local geometric information may not be easily preserved in a specific ordered list. 

More recently, various approaches have been proposed for effective local feature learning. PointCNN \cite{li2018pointcnn} aligns points in a certain order by predicting a transformation matrix for local point set, which inevitably leads to sensitivity in point order since the operation is not permutation-invariant. SpiderCNN \cite{xu2018spidercnn} defines its convolution kernel as a family of polynomial functions, relying on the neighbors' order. PCNN \cite{atzmon2018point} designs point kernels based on the spatial coordinates and further KPConv \cite{thomas2019kpconv} presents a scalable convolution using explicit kernel points.
RS-CNN \cite{liu2019relation} assigns channel-wise weights to neighboring point features according to the geometric relations learned from 10-D vectors. ShellNet \cite{zhang2019shellnet} splits local point set into several shell areas, from which features are extracted and aggregated.
Recently, \cite{zhao2020point,guo2020pct} utilize the successful transformer structures in natural language processing \cite{vaswani2017attention,wu2019pay} to build dense self-attention between local and global features.

\begin{figure*}
	\centering
	\includegraphics[width=0.95\linewidth]{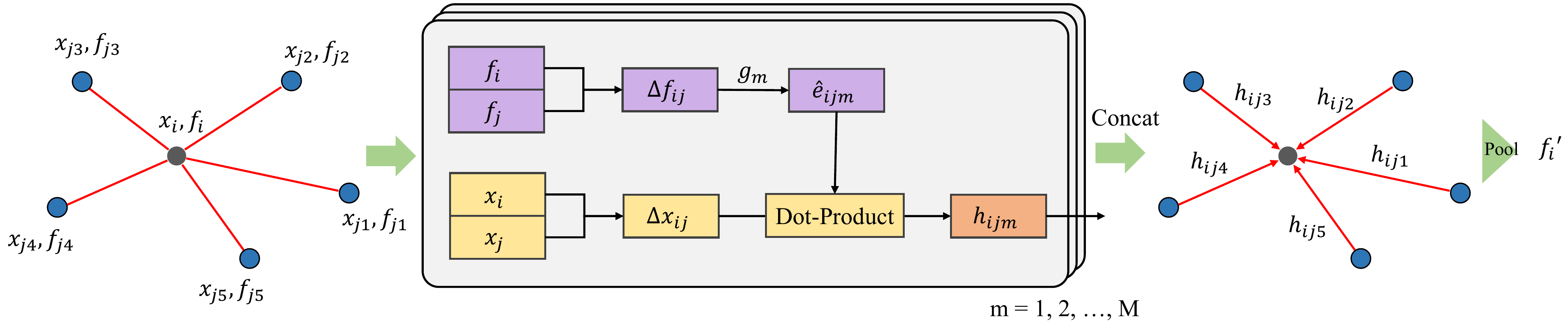}
	\caption{The illustration of AdaptConv processed in the neighborhood of a target point $x_i$. An adaptive kernel $\hat{e}_{ijm}$ is generated from the feature input $\Delta f_{ij}$ of a pair of points on the edge, which is then convolved with the corresponding spatial input $\Delta x_{ij}$. Concatenating $h_{ijm}$ of all dimensions yields the edge feature $h_{ij}$. Finally, the output feature $f_i'$ of the central point is obtained through a pooling function. AdaptConv differs from other graph convolutions in that the convolution kernel is unique for each pair of points.}
	\label{fig:kernel}
\end{figure*}

The graph-based methods treat points as nodes of a graph, and establish edges according to their spatial/feature relationships. Graph is a natural representation for a point cloud to model local geometric structures but is challenging for processing due to its irregularity. 
The notion of Graph Convolutional Network is proposed by \cite{kipf2016semi}, which generalizes convolution operations over graphs by averaging features of adjacent nodes. Similar ideas \cite{shen2018mining,wang2019dynamic,hua2018pointwise,li2018pointcnn,lei2020spherical} have been explored to extract local geometric features from local points. 
Shen et al. \cite{shen2018mining} define kernels according to euclidean distances and geometric affinities in the neighboring points. DGCNN \cite{wang2019dynamic} gathers nearest neighboring points in the feature space, followed by the EdgeConv operators for feature extraction, in order to identify semantic cues dynamically. MoNet \cite{monti2017geometric} defines the convolution as Gaussian mixture models in a local pseudo-coordinate system. Inspired by the idea of attention mechanism, several works \cite{velivckovic2017graph, wang2019graph, verma2018feastnet} propose to assign proper attentional weights to different points/filters. 3D-GCN \cite{lin2020convolution} develops deformable kernels, focusing on shift and scale-invariant properties in point cloud analysis.

%RS-CNN \cite{liu2019relation} assigns channel-wise weights to neighboring point features according to the geometric relations learned from 10-D vectors. ShellNet \cite{zhang2019shellnet} splits local point set into several shell area, from which features are extracted and aggregated. SPH3D \cite{lei2020spherical} presents a deformed version of traditional 3D CNN, namely using discrete spherical kernels in 3D space. 

\textbf{Convolution on point clouds.}
State-of-the-art researches have proposed many methods to define a proper convolution on point clouds. To improve the basic designs using fixed MLPs in PointNet/PointNet++, a variety of works \cite{velivckovic2017graph, wang2019graph, verma2018feastnet,thomas2019kpconv,liu2019relation} try to introduce weights based on the learned features, with more varients of convolution inputs \cite{wang2019dynamic,monti2017geometric,xu2018spidercnn}. 
%However, the limitation caused by the isotropy of convolution still exists for those relying on fixed kernels. 
Other methods \cite{simonovsky2017dynamic,wu2019pointconv,jia2016dynamic} try to learn a dynamic weight for the convolution. However, their idea is to approximate weight functions from the direct 3D coordinates while AdaptConv uses features to learn the kernels, which represents more adaptiveness. In addition, their implementation is heavily memory consuming when convolving with high-dimensional features. Thus, the main focus of this paper is to handle the isotropy of point cloud convolutions, by developing an adaptive kernel that is unique to each point in the convolution.
%However, due to the irrgularity of point data, there does not exist an explicit mapping between the neighbors and convolution kernels, unlike the cases of CNNs for image inputs. Thus, they all use a fixed kernel shared among points in the point cloud. In this paper, the proposed AdaptConv is designed to handle the isotropy of graph convolutions, which develop adaptive kernels that is unique to points in the convolution.

\label{sec:related}

\begin{figure*}
	\centering
	\includegraphics[width=0.95\linewidth]{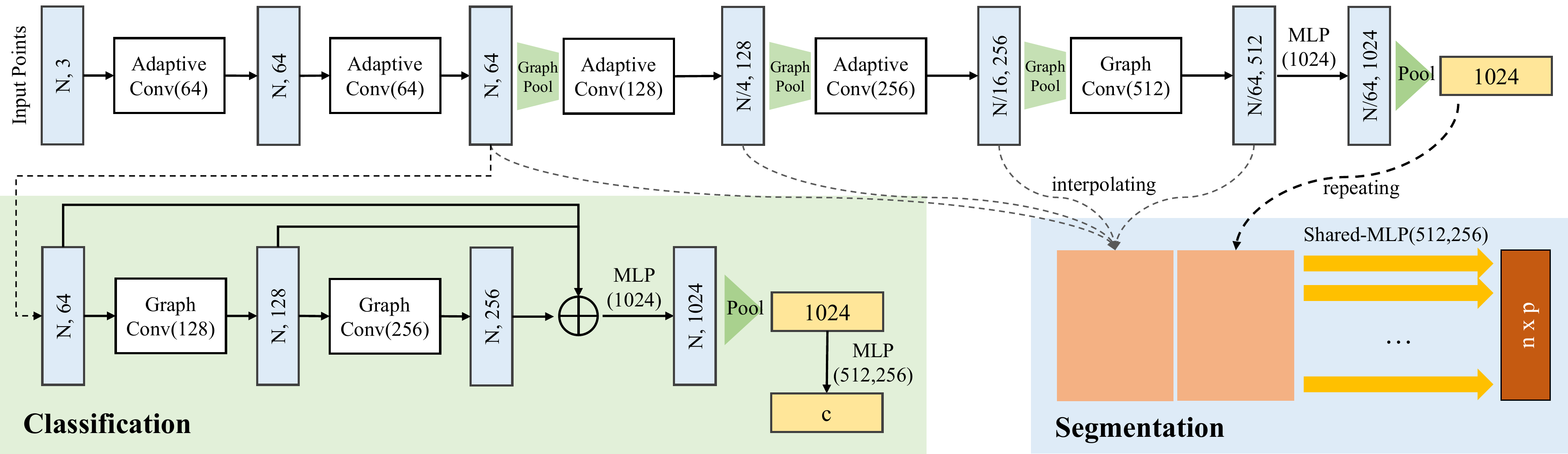}
	
	\caption{AdaptConv network architectures for classification and segmentation tasks. The GraphConv layer denotes our standard convolution without an adaptive kernel. The segmentation model uses pooling and interpolating to build a hierachical graph structure, while the classification model applies a dynamic structure \cite{wang2019dynamic}.}
	\label{fig:architecture}
\end{figure*}

\section{Method}
We exploit local geometric characteristics in point cloud analysis by proposing a novel adaptive graph convolution (AdaptConv) in the spirit of graph neural networks (Sec.~\ref{sec:method:adapt}). Afterwards, we discuss several choices for the feature decisions in the adaptive convolution (Sec.~\ref{sec:method:feature}). The details of the constructed networks are shown in Sec.~\ref{sec:method:architecture}. %Finally, we compare our model with prior graph convolutions in Sec.~\ref{sec:method:cmp}.

\subsection{Adaptive graph convolution}
\label{sec:method:adapt}
%
%For the sake of clarity, 
We denote the input point cloud as $\mathcal{X} = \{x_i | i=1,2,...,N\} \in \mathbb{R}^{N \times 3}$ with the corresponding features defined as $\mathcal{F} = \{f_i | i=1,2,...,N\} \in \mathbb{R}^{N \times D}$. 
Here, $x_i$ processes the $(\mathbf{x},\mathbf{y},\mathbf{z})$ coordinates of the i-th point, and, in other cases, can be potentially combined with a vector of additional attributes, such as normal and color. 
%
%To start with, 
We then compute a directed graph $\mathcal{G}(\mathcal{V},\mathcal{E})$ from the given point cloud where $\mathcal{V} = \{1,...,N\}$ and $\mathcal{E} \subseteq \mathcal{V} \times \mathcal{V}$ represents the set of edges. 
We construct the graph by employing the $k$-nearest neighbors (KNN) of each point including self-loop. 
Given the input $D$-dimensional features, our AdaptConv layer is designed to produce a new set of $M$-dimensional features with the same number of points while attempting to more accurately reflect local geometric characteristics than previous graph convolutions.

Denote that $x_i$ is the central point in the graph convolution, and $\mathcal{N}(i) = \{j : (i,j) \in \mathcal{E}\}$ is a set of point indices in its neighborhood. 
%
%In order to capture local geometric information within the neighboring patch, 
Due to the irregularity of point clouds, previous methods usually apply a fixed kernel function on all neighbors of $x_i$ to capture the geometric information of the patch.
However, different neighbors may reflect different feature correspondences with $x_i$, particularly when $x_i$ is located at salient regions, such as corners or edges.
In this regard, the fixed kernel may incapacitate the geometric representations generated from the graph convolution for classification and, particularly, segmentation.
%each pair of $(x_i, x_j)$ to represent its geometric relationship
%, while the relationships on edges are dynamic throughout the point cloud. 
%

In contrast, we endeavor to design an adaptive kernel to capture the distinctive relationships between each pair of points. 
To achieve this, for each channel in the output $M$-dimensional feature, our AdaptConv dynamically generates a kernel using a function over the point features $(f_i, f_j)$:
\begin{equation}
\hat{e}_{ijm} = g_m(\Delta f_{ij}), j \in \mathcal{N}(i).
\end{equation}
Here, $m = 1,2,...,M$ indicates one of the $M$ output dimensions corresponding to a single filter defined in our AdaptConv. 
In order to combine the global shape structure and feature differences captured in a local neighborhood \cite{wang2019dynamic}, we define $\Delta f_{ij} = [f_i, f_j-f_i]$ as the input feature for the adaptive kernel, where $[\cdot,\cdot]$ is the concatenation operation. 
The $g(\cdot)$ is a feature mapping function, and here we use a multilayer perceptron. 

Like the computations in 2D convolutions, which obtain one of the $M$ output dimensions by convolving the $D$ input channels with the corresponding filter weights, our adaptive kernel is convolved with the corresponding points $(x_i, x_j)$:
\begin{equation}
h_{ijm} = \sigma \left \langle \hat{e}_{ijm}, \Delta x_{ij} \right \rangle, \label{equ:convolution}
\end{equation}
where $\Delta x_{ij}$ is defined as $[x_i, x_j-x_i]$ similarly, $\langle \cdot, \cdot \rangle$ represents the inner product of two vectors outputting $h_{ijm} \in \mathbb{R}$ and $\sigma$ is a nonlinear activation function. As shown in Fig.~\ref{fig:kernel} (middle part), the m-th adaptive kernel $\hat{e}_{ijm}$ is combined with the spatial relations $\Delta x_{ij}$ of the corresponding point $x_j \in \mathbb{R}^3$, which means the size of the kernel should be matched in the dot product, \ie, the aforementioned feature mapping is $g_m: \mathbb{R}^{2D} \rightarrow \mathbb{R}^{6}$. In this way, the spatial positions in the input space can be efficiently incorporated into each layer, combined with the feature correspondences extracted dynamically from our kernel. Stacking the $h_{ijm}$ of each channel yields the edge feature $h_{ij} = [h_{ij1}, h_{ij2}, ..., h_{ijM}] \in \mathbb{R}^M$ between the connected points $(x_i, x_j)$.
Finally, we define the output feature of the central point $x_i$ by applying an aggregating function over all the edge features in the neighborhood (see Fig.~\ref{fig:kernel} (right part)):
\begin{equation}
f_i' = \max_{j \in \mathcal{N}(i)} h_{ij},
\end{equation}
where $\max$ is a channel-wise max-pooling function.
Overall, the convolution weights of AdaptConv are defined as $\Theta = (g_1, g_2, ..., g_M)$.

\subsection{Feature decisions}
\label{sec:method:feature}
In our method, AdaptConv generates an adaptive kernel for each pair of points according to their individual features $(f_i, f_j)$. Then, the kernel $\hat{e}_{ijm}$ is applied to the point pair of $(x_i, x_j)$ in order to describe their spatial relations in the input space. The feature decision of $\Delta x_{ij}$ in the convolution of Eq.~\ref{equ:convolution} is an important design. In other cases, the inputs can be $x_i \in \mathbb{R}^E$ including additional dimensions representing other valuable point attributes, such as point normals and colors. 
%Prior works treat the multi-feature input as a whole, using a general mapping over concatenated features from distinct fields. 
By modifying the adaptive kernel to $g_m: \mathbb{R}^{2D} \rightarrow \mathbb{R}^{2E}$, our AdaptConv can also capture the relationships between feature dimensions and spatial coordinates which are from different domains. Note that, this is another option in our AdaptConv design, and we use the spatial positions as input $x_i$ by default in the convolution in our experiments.

As an optional choice, we replace the $\Delta x_{ij}$ with $\Delta f_{ij}$ in Eq.~\ref{equ:convolution} with a modified dimension of $\hat{e}_{ijm}$. Therefore, the adaptive kernel of a pair of points is designed to establish the relations of their current features $(f_i, f_j)$ in each layer. This is a more direct solution, similar to other convolution operators, that produces a new set of learned features from features in the preceding layer of the network.
However, we recommend xyz rather than feature in that: (i) the point feature $f_j$ has been already included in the adaptive kernel and convolving again with $f_j$ leads to redundancy of feature information; (ii) it is easier to learn spatial relations through MLPs, instead of detecting feature correspondences in a high-dimensional space (\eg 64, 128 dimensional features); (iii) the last reason is the memory cost and more specifically the large computational graph in the training stage which cannot be avoided. We evaluate all these choices in Sec.~\ref{sec:eval:ablation}. 
%However, as convolving the generated kernel with high-dimensional features introduces extensive cost of memory usage, it is not recommended compared with a more efficient version of using point coordinates.  We evaluate all these choices in Sec.~\ref{sec:eval:ablation}. 
%In general, by utilizing the idea of adaptive kernel, the convolution on point clouds can be more flexible, adapting to different features and structures.

\subsection{Network architecture}
\label{sec:method:architecture}
We design two network architectures for point cloud classification and segmentation tasks using the proposed AdaptConv layer. The network architectures are shown in Fig.~\ref{fig:architecture}. In our experiments, the AdaptConv kernel function is implemented as a two-layer MLP with residual connections to extract important geometric information. More details are available in the supplemental material. The standard graph convolution layer with a fixed kernel uses the same feature inputs $\Delta f_{ij}$ as in the adaptive kernels.
%For both tasks, we apply two layers of adaptive convolution, followed by another two standard graph convolutions for feature processing from input points. This combination achieves pleasing results while reducing the model  size as shown in Sec.~\ref{sec:eval:eff}. The structure of standard graph convolution is described in Fig.~\ref{fig:architecture} (left bottom part), which utilizes a shared function (fixed kernel) to extract edge features.The feature inputs $\Delta f_{ij}$ are the same as in our adaptive kernel design.

%\textbf{Dynamic graph update.} Following \cite{wang2019dynamic}, we update the graph structure in each layer according to the feature similarity among points, rather than fixed using spatial positions. That is, in each layer, the edge set $\mathcal{E}^{(l)}$ is recomputed where the neighborhood of point $x_i$ is $\mathcal{N}(i) = \{j_{i_1}, j_{i_2}, ..., j_{i_k}\}$ such that the corresponding features $f_{j_{i_1}}, f_{j_{i_2}}, ..., f_{j_{i_k}}$ are closest to $f_i$. This encourages the network to organize the graph semantically, grouping together similar points in the feature space but not solely considering their proximity in the spatial inputs. Thus, the receptive field of local points is expanded, leading to a propagation of local information throughout the point cloud. 
%Note that, in the convolution with adaptive kernel in Eq.~\ref{equ:convolution}, the $\Delta x_{ij}$ corresponds to the feature pair $(f_i, f_j)$ while may not be spatially close.

\textbf{Graph pooling.} For segmentation tasks, we reduce the number of points progressively in order to build the network in a hierarchical architecture. The point cloud is subsampled using furthest point sampling algorithm \cite{QiSMG17} with a sampling rate of 4, and is applied by a pooling layer to output aggregated features on the coarsened graph. In each graph pooling layer, a new graph is constructed corresponding to the sampled points. The feature pooled at each point in the subcloud can be simply obtained by a max-pooling function within its neighborhood. Alternatively, we can use a AdaptConv layer to aggregate this pooled features.
To predict point-wise labels for segmentation purpose, we need to interpolate deeper features from subsampled cloud to the original points. Here, we use the nearest upsamping to get the features for each layer, which are concatenated for the final point-wise features.

\textbf{Segmentation network.} Our segmentation network architecture is illustrated in Fig.~\ref{fig:architecture}. The AdaptConv encoder includes 5 layers of convolutions in which the last one is a standard graph convolution layer, as well as several graph pooling layers. The subsampled features are interpolated and concatenated for the final point features which are fed to the decoder part. 

\textbf{Classification network.} The classification network uses a similar encoder part as in the segmentation model (see Fig.~\ref{fig:architecture}). For sparser point clouds used in the ModelNet40 classification dataset, we simply apply dynamic graph structures \cite{wang2019dynamic} without pooling and interpolation. Specifically, the graph structure is updated in each layer according to the feature similarity among points, rather than fixed using spatial positions. That is, in each layer, the edge set $\mathcal{E}_l$ is recomputed where the neighborhood of point $x_i$ is $\mathcal{N}(i) = \{j_1, j_2, ..., j_k\}$ such that the corresponding features $f_{j_1}, f_{j_2}, ..., f_{j_k}$ are closest to $f_i$. This encourages the network to organize the graph semantically and expands the receptive field of local neighborhood by grouping together similar points in the feature space.
\label{sec:method}

\section{Evaluation}
In this section, we evaluate our models using AdaptConv for point cloud classification, part segmentation and indoor segmentation tasks. Detailed network architectures and comparisons are provided.

\begin{table}[t]
	\centering
	%\small
	\footnotesize
	\begin{tabular}{c|cccc} %表格6列 全部居中显示
		\toprule[1pt]
		Method & Input & \#points & mAcc(\%) & OA(\%) \\
		\midrule[0.3pt]
		\midrule[0.3pt]
		3DShapeNetParts \cite{wu20153d}						& voxel & - & 77.3 & 84.7 \\
		VoxNet \cite{maturana2015voxnet}				& voxel & - & 83.0 & 85.9 \\
		Subvolume \cite{qi2016volumetric}				& voxel & - & 86.0 & 89.2 \\
		\midrule[0.3pt]
		PointNet \cite{QiSMG17}							& xyz & 1k & 86.0 & 89.2 \\
		PointNet++ \cite{qi2017pointnet++}				& xyz, normal & 5k & - & 91.9 \\
		Kd-Net \cite{klokov2017escape}					& xyz & 1k & - & 90.6 \\
		SpecGCN \cite{wang2018local}					& xyz & 1k & - & 92.1 \\
		SpiderCNN \cite{xu2018spidercnn}				& xyz, normal & 5k & - & 92.4 \\
		PointCNN \cite{li2018pointcnn}					& xyz & 1k & 88.1 & 92.2 \\
		SO-Net \cite{li2018so}							& xyz, normal & 5k & - & \textbf{93.4} \\
		DGCNN \cite{wang2019dynamic}					& xyz & 1k & 90.2 & 92.9 \\
		KPConv \cite{thomas2019kpconv}					& xyz & 6.8k & - & 92.9 \\
		3D-GCN \cite{lin2020convolution}				& xyz & 1k & - & 92.1 \\
		PointASNL \cite{yan2020pointasnl}				& xyz, normal & 1k & - & 93.2 \\
		\midrule[0.3pt]
		Ours											& xyz & 1k & \textbf{90.7} & \textbf{93.4} \\
		\bottomrule[1pt]
	\end{tabular}
	\vspace{5pt}
	\caption{Classification results on ModelNet40 dataset. Our network achieves the best results according to the mean class accuracy (mAcc) and overall accuracy (OA).}
	\label{table:cls_results}
\end{table}

\begin{table*}
	\centering
	%\small
	\footnotesize
	%\scriptsize
	\setlength{\tabcolsep}{1mm}
	\begin{tabular}{c|cc|cccccccccccccccc} %表格6列 全部居中显示
		\toprule[1pt]
		Method & mcIoU & mIoU & air & bag & cap & car & chair & ear & guitar & knife & lamp & laptop & motor & mug & pistol & rocket & skate & table  \\
		& & & 					plane &  &  	&	  &	     & phone &      &		&		&	 	& bike & 		&		&		 & board &	\\
		\midrule[0.3pt]
		Kd-Net \cite{klokov2017escape} 	& 77.4 & 82.3 & 80.1 & 74.6 & 74.3 & 70.3 & 88.6 & 73.5 &90.2 & 87.2 & 81.0 & 94.9 & 87.4 & 86.7 & 78.1 & 51.8 & 69.9 & 80.3 \\
		%MRTNet \cite{gadelha2018multiresolution} & 79.3 & 83.0 & 81.0 & 76.7 & 87.0 & 73.8 & 89.1 & 67.6 & 90.6 & 85.4 & 80.6 & 95.1 & 64.4 & 91.8 & 79.7 & 87.0 & 69.1 & 80.6 \\
		PointNet \cite{QiSMG17}	& 80.4 & 83.7 & 83.4 & 78.7 & 82.5 & 74.9 & 89.6 & 73.0 & 91.5 & 85.9 & 80.8 & 95.3 & 65.2 & 93.0 & 81.2 & 57.9 & 72.8 & 80.6 \\
		PointNet++ \cite{qi2017pointnet++} & 81.9 & 85.1 & 82.4 & 79.0 & 87.7 & 77.3 & 90.8 & 71.8 & 91.0 & 85.9 & 83.7 & 95.3 & 71.6 & 94.1 & 81.3 & 58.7 & 76.4 & 82.6\\
		SO-Net \cite{li2018so} & 81.0 & 84.9 & 82.8 & 77.8 & 88.0 & 77.3 & 90.6 & 73.5 &90.7 & 83.9 & 82.8 & 94.8 & 69.1 & 94.2 & 80.9 & 53.1 & 72.9 & 83.0 \\
		DGCNN \cite{wang2019dynamic} & 82.3 & 85.2 & 84.0 & 83.4 & 86.7 & 77.8 & 90.6 & 74.7 & 91.2 & 87.5 & 82.8 & 95.7 & 66.3 & 94.9 & 81.1 &    63.5 & 74.5 & 82.6\\
		%PCNN \cite{atzmon2018point} & - & 85.1 & 82.4 & 80.1 & 85.5 & 79.5 & 90.8 & 73.2 & 91.3 & 86.0 & 85.0 & 95.7 & 73.2 & 94.8 & 83.3 & 51.0 & 75.0 & 81.8 \\
		PointCNN \cite{li2018pointcnn} & - & 86.1 & 84.1 & 86.4 & 86.0 & 80.8 & 90.6 & 79.7 & 92.3 & 88.4 & 85.3 & 96.1 & 77.2 & 95.3 & 84.2 & 64.2 & 80.0 & 83.0 \\
		PointASNL \cite{yan2020pointasnl} & - & 86.1 & 84.1 & 84.7 & 87.9 & 79.7 & 92.2 & 73.7 & 91.0 & 87.2 & 84.2 & 95.8 & 74.4 & 95.2 & 81.0 & 63.0 & 76.3 & 83.2\\
		3D-GCN \cite{lin2020convolution} & 82.1 & 85.1 & 83.1 & 84.0 & 86.6 & 77.5 & 90.3 & 74.1 & 90.9 & 86.4 & 83.8 & 95.6 & 66.8 & 94.8 & 81.3 & 59.6 & 75.7 & 82.8 \\
		KPConv \cite{thomas2019kpconv} & \textbf{85.1} & \textbf{86.4} & 84.6 & 86.3 & 87.2 & 81.1 & 91.1 & 77.8 & 92.6 & 88.4 & 82.7 & 96.2 & 78.1 & 95.8 & 85.4 & 69.0 & 82.0 & 83.6\\
		\midrule[0.3pt]
		Ours  & 83.4 & \textbf{86.4} & 84.8 & 81.2 & 85.7 & 79.7 & 91.2 & 80.9 & 91.9 & 88.6 & 84.8 & 96.2 & 70.7 & 94.9 & 82.3 & 61.0 & 75.9 & 84.2 \\
		\bottomrule[1pt]
	\end{tabular}
	\vspace{5pt}
	\caption{Part segmentation results on ShapeNetPart dataset evaluated as  the mean class IoU (mcIoU) and mean instance IoU (mIoU).}
	\label{table:partseg_results}
\end{table*}

\subsection{Classifcation}
\textbf{Data.} We evaluate our model on ModelNet40 \cite{wu20153d} dataset for point cloud classification. This dataset contains 12,311 meshed CAD models from 40 categories, where 9,843 models are used for training and 2,468 models for testing. We follow the experimental setting of \cite{QiSMG17}. 1024 points are uniformly sampled for each object and we only use the $(\mathbf{x}, \mathbf{y}, \mathbf{z})$ coordinates of the sampled points as input. The data augmentation procedure includes shifting, scaling and perturbing of the points.

\textbf{Network configuration.} The network architecture is shown in Fig.~\ref{fig:architecture}. Following \cite{wang2019dynamic}, we recompute the graph based on the feature similarity in each layer. The number $k$ of neighborhood size is set to 20 for all layers. Shortcut connections are included and one shared fully-connected layer (1024) is applied to aggregate the multi-scale features. The global feature is obtained using a max-pooling function. All layers are with LeakyReLU and batch normalization.
%Afterwards, we collect features from all layers and apply one shared fully-connected layer (1024). Then, a max pooling function is applied to get the global feature for the point cloud, after which fully-connected layers (512, 256) are used for the classification output. In the last two fully-connected layers, we use dropout with a keep probability of 0.5. All layers are with LeakyReLU and batch normalization.
We use SGD optimizer with momentum set to 0.9. The initial learning rate is 0.1 and is dropped until 0.001 using cosine annealing \cite{loshchilov2016sgdr}. The batch size is set to 32 for all training models. We use PyTorch \cite{paszke2019pytorch} implementation and train the network on a RTX 2080 Ti GPU. The hyperparameters are chosen in a similar way for other tasks.

\textbf{Results.} We show the results for classification  in Tab.~\ref{table:cls_results}. The evaluation metrices on this dataset are the mean class accuracy (mAcc) and the overall accuracy (OA). Our model achieves the best scores on this dataset. For a clear comparison, we show the input data types and the number of points corresponding to each method. Our AdaptConv only considers the point coordinates as input with a relatively small size of 1k points, which already outperforms other methods using larger inputs.

\subsection{Part segmentation}
\textbf{Data.} We further test our model for part segmentation task on ShapeNetPart dataset \cite{yi2016scalable}. This dataset contains 16,881 shapes from 16 categories, with 14,006 for training and 2,874 for testing. Each point is annotated with one label from 50 parts and each point cloud contains 2-6 parts. We follow the experimental setting of \cite{qi2017pointnet++} and use their provided data for benchmarking purpose. 2,048 points are sampled from each shape. The input attributes include the point normals apart from the 3D coordinates. 

\textbf{Network configuration.} Following \cite{QiSMG17}, we include a one-hot vector representing category types for each point. It is stacked with the point-wise features to compute the segmentation results. Other training parameters are set the same as in our classification task. Note that, we use spatial positions (without normals) as $\Delta x_{ij}$ as discussed in Sec.~\ref{sec:method:feature}. Other choices will be evaluated later in Sec.~\ref{sec:eval:ablation}.

\textbf{Results.} We report the mean class IoU (mcIoU) and mean instance IoU (mIoU) in Tab.~\ref{table:partseg_results}. Following the evaluation scheme of \cite{QiSMG17}, the IoU of a shape is computed by averaging the IoU of each part. The mean IoU (mIoU) is computed by averaging the IoUs of all testing instances. The class IoU (mcIoU) is the mean IoU over all shape categories. We also show the class-wise segmentation results. Our model achieves the state-of-the-art performance compared with other methods.

\begin{table}[t]
	\centering
	\small
	%\footnotesize
	\setlength{\tabcolsep}{3.5mm}
	\begin{tabular}{c|cc} %表格6列 全部居中显示
		\toprule[1pt]
		Ablations & mcIoU(\%) & mIoU(\%) \\
		\midrule[0.3pt]
		\midrule[0.3pt]
		GraphConv							& 81.9 & 85.5 \\
		Attention Point						& 78.0 & 83.3 \\
		Attention Channel					& 77.9 & 83.0\\
		\midrule[0.3pt]
		Feature								& 82.2 & 85.9 \\
		Normal								& 83.2 & 86.2 \\
		Initial attributes					& 83.2 & 86.1 \\
		Ours								& \textbf{83.4} & \textbf{86.4} \\
		\bottomrule[1pt]
	\end{tabular}
	\vspace{5pt}
	\caption{Ablation studies on ShapeNetPart dataset for part segmentation.}
	\label{table:ab:seg}
\end{table}

\begin{table*}
	\centering
	%\small
	\footnotesize
	%\scriptsize
	\setlength{\tabcolsep}{1mm}
	\begin{tabular}{c|ccc|ccccccccccccc} %表格6列 全部居中显示
		\toprule[1pt]
		Method & OA & mAcc & mIoU & ceiling & floor & wall & beam & column & window & door & table & chair & sofa & bookcase & board & clutter  \\
		\midrule[0.3pt]
		PointNet \cite{QiSMG17} & – & 49.0 & 41.1 & 88.8 & 97.3 & 69.8 & 0.1 & 3.9 & 46.3 & 10.8 & 59.0 & 52.6 & 5.9 & 40.3 & 26.4 & 33.2 \\
		SegCloud \cite{tchapmi2017segcloud} & – & 57.4 & 48.9& 90.1& 96.1& 69.9& 0.0& 18.4& 38.4& 23.1& 70.4& 75.9& 40.9& 58.4& 13.0& 41.6 \\
		PointCNN \cite{li2018pointcnn} &85.9& 63.9& 57.3& 92.3& 98.2& 79.4& 0.0& 17.6& 22.8& 62.1& 74.4& 80.6& 31.7& 66.7& 62.1& 56.7 \\
		PCCN \cite{wang2018deep} &–& 67.0& 58.3& 92.3& 96.2& 75.9& 0.3& 6.0& 69.5& 63.5& 66.9& 65.6& 47.3& 68.9& 59.1& 46.2 \\
		PointWeb \cite{zhao2019pointweb} & 87.0& 66.6& 60.3& 92.0& 98.5& 79.4& 0.0& 21.1& 59.7& 34.8& 76.3& 88.3& 46.9& 69.3& 64.9& 52.5 \\
		HPEIN \cite{jiang2019hierarchical} & 87.2& 68.3& 61.9& 91.5& 98.2& 81.4& 0.0& 23.3& 65.3& 40.0& 75.5& 87.7& 58.5& 67.8& 65.6& 49.4 \\
		GAC \cite{wang2019graph} &87.7 &-& 62.8& 92.2& 98.2& 81.9& 0.0& 20.3& 59.0& 40.8& 78.5& 85.8& 61.7& 70.7& 74.6& 52.8 \\
		KPConv \cite{thomas2019kpconv} &– &72.8& 67.1& 92.8& 97.3& 82.4& 0.0& 23.9& 58.0& 69.0& 81.5& 91.0& 75.4& 75.3& 66.7& 58.9 \\
		PointASNL \cite{yan2020pointasnl} &87.7& 68.5& 62.6& 94.3& 98.4& 79.1& 0.0& 26.7& 55.2& 66.2& 83.3& 86.8& 47.6& 68.3& 56.4& 52.1 \\
		\midrule[0.3pt]
		Ours &\textbf{90.0}& \textbf{73.2}&\textbf{67.9} & 93.9& 98.4& 82.2&  0.0& 23.9& 59.1& 71.3& 91.5& 81.2& 75.5& 74.9& 72.1& 58.6 \\ 
		\bottomrule[1pt]
	\end{tabular}
	\vspace{5pt}
	\caption{Semantic segmentation results on S3DIS dataset evaluated on Area 5. We report the mean classwise IoU (mIoU), mean classwise accuracy (mAcc) and overall accuracy (OA). IoU of each class is also provided.}
	\label{table:scene_results}
\end{table*}

\subsection{Indoor scene segmentation}
\textbf{Data.} Our third experiment shows the semantic segmentation performance of our model on the S3DIS dataset \cite{armeni20163d}. This dataset contains 3D RGB point clouds from six indoor areas of three different buildings, covering a total of 271 rooms. Each point is annotated with one semantic label from 13 categories. For a common evaluation protocol \cite{tchapmi2017segcloud,QiSMG17,landrieu2018large}, we choose Area 5 as the test set which is not in the same building as other areas.

\textbf{Real scene segmentation.} The large-scale indoor datasets reveal more challenges, covering larger scenes in a real-world enviroment with a lot more noise and outliners. Thus, we follow the experimental settings of KPConv \cite{thomas2019kpconv}, and train the network using randomly sampled clouds in spheres. The subclouds contain more points with varing sizes, and are stacked into batches for training. In the test stage, spheres are uniformly picked in the scenes, and we ensure each point is tested several times using a voting scheme. The input point attributes include the RGB colors and the original heights.

\begin{figure}[t]
	\newlength{\unit}
	\setlength{\unit}{0.27\linewidth}
	\centering
	
	\includegraphics[width=\unit]{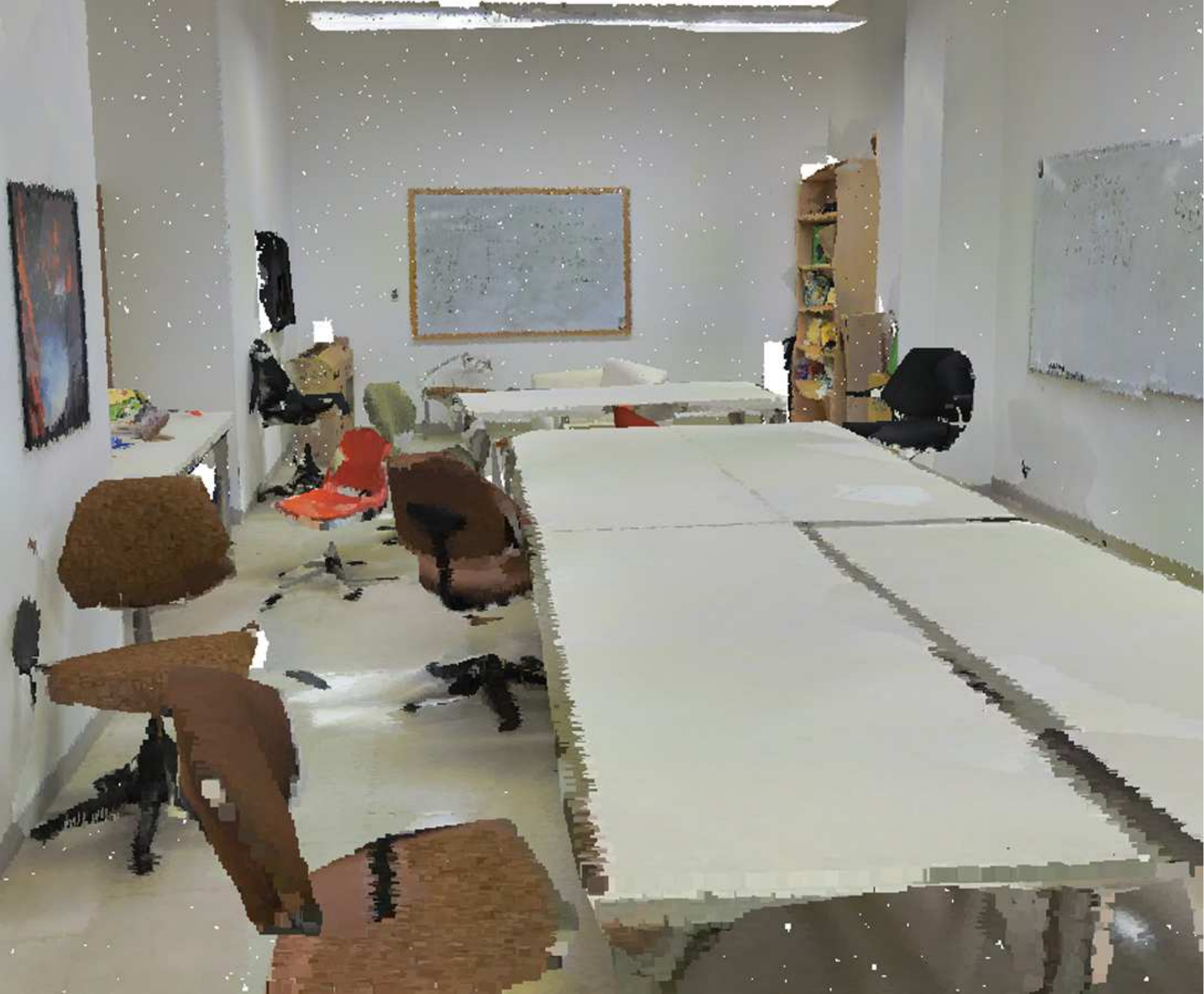}%
	\hspace{2pt}\includegraphics[width=\unit]{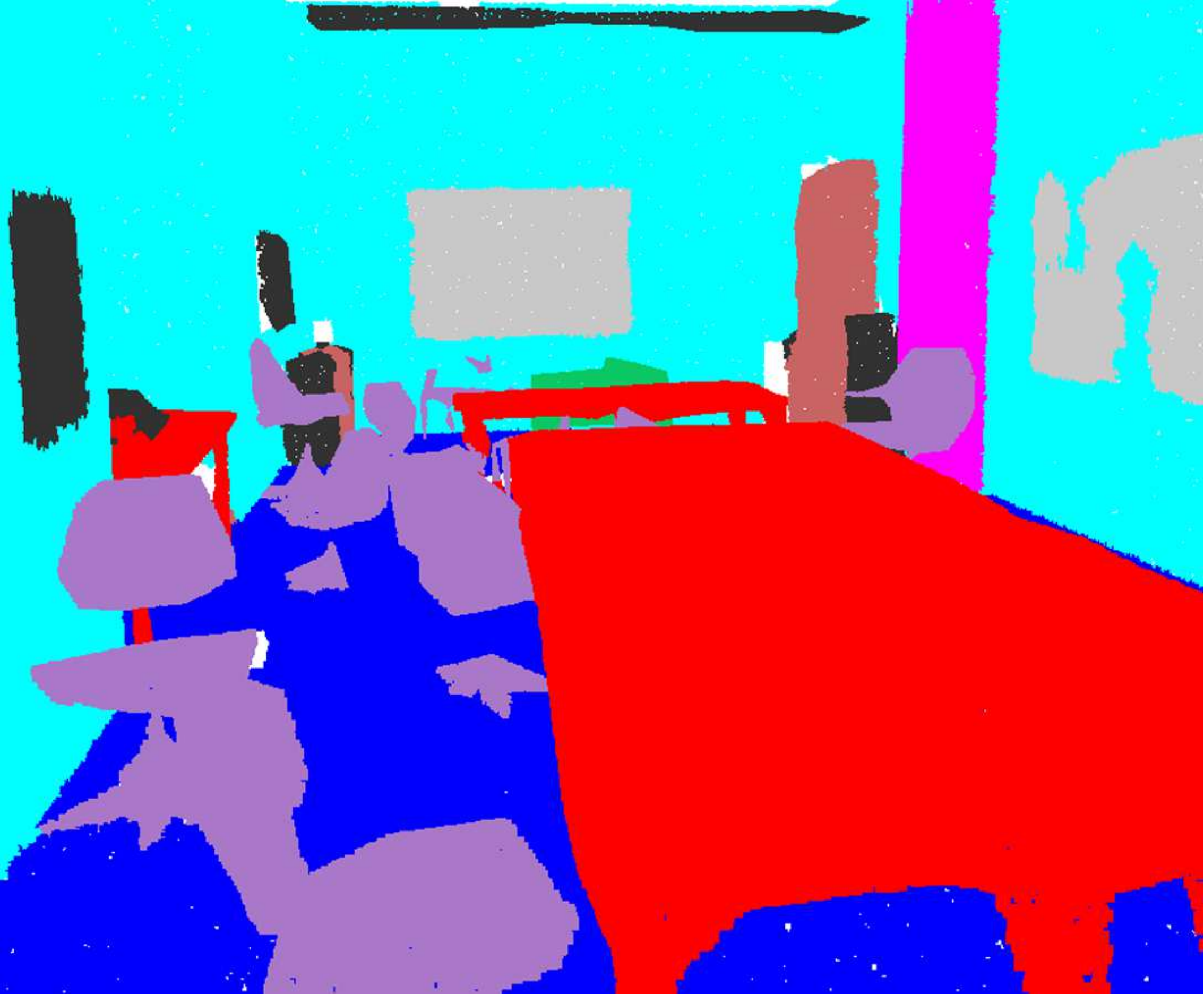}%
	\hspace{2pt}\includegraphics[width=\unit]{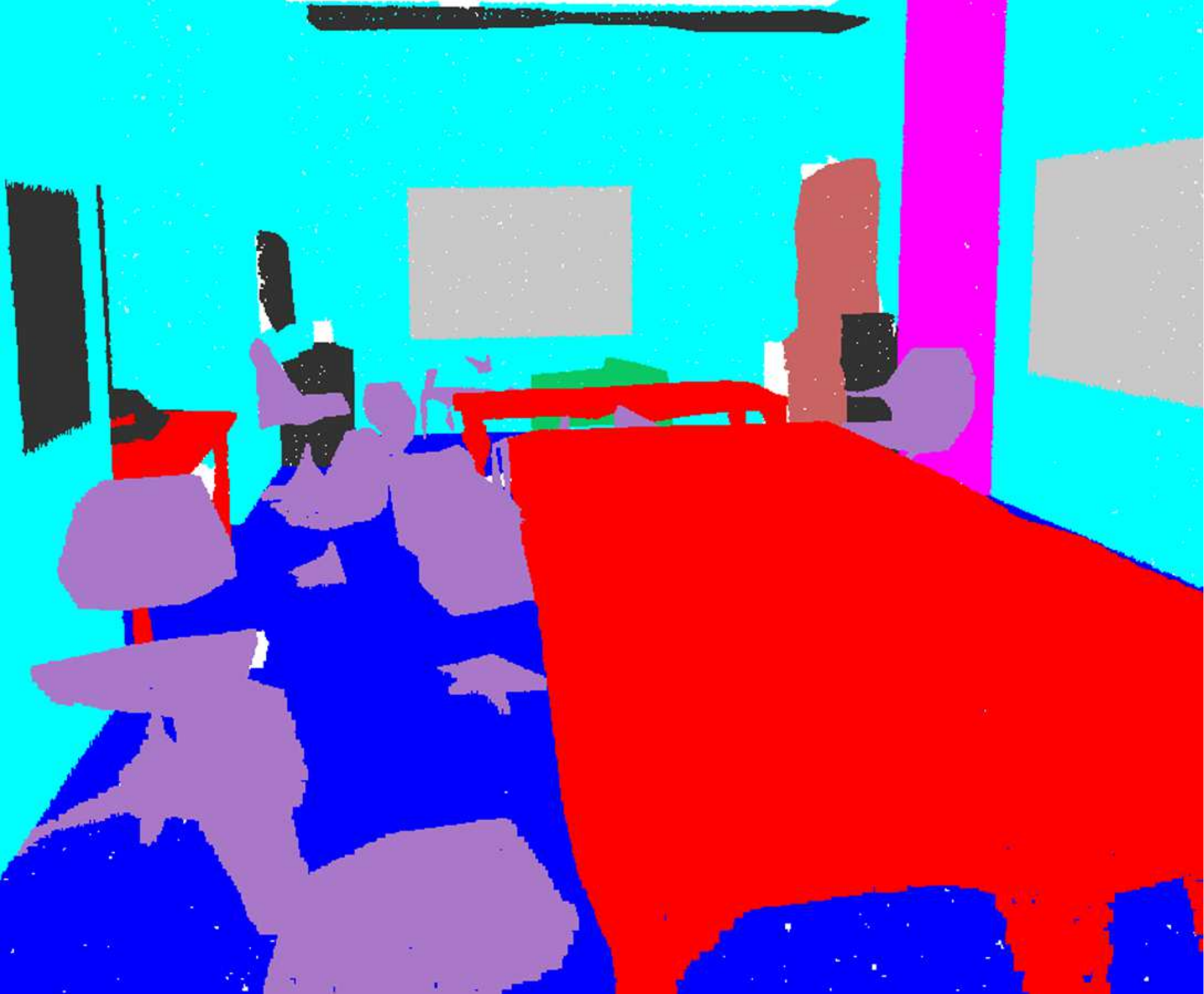}
	
	\vspace{3pt}
	
	\includegraphics[width=\unit]{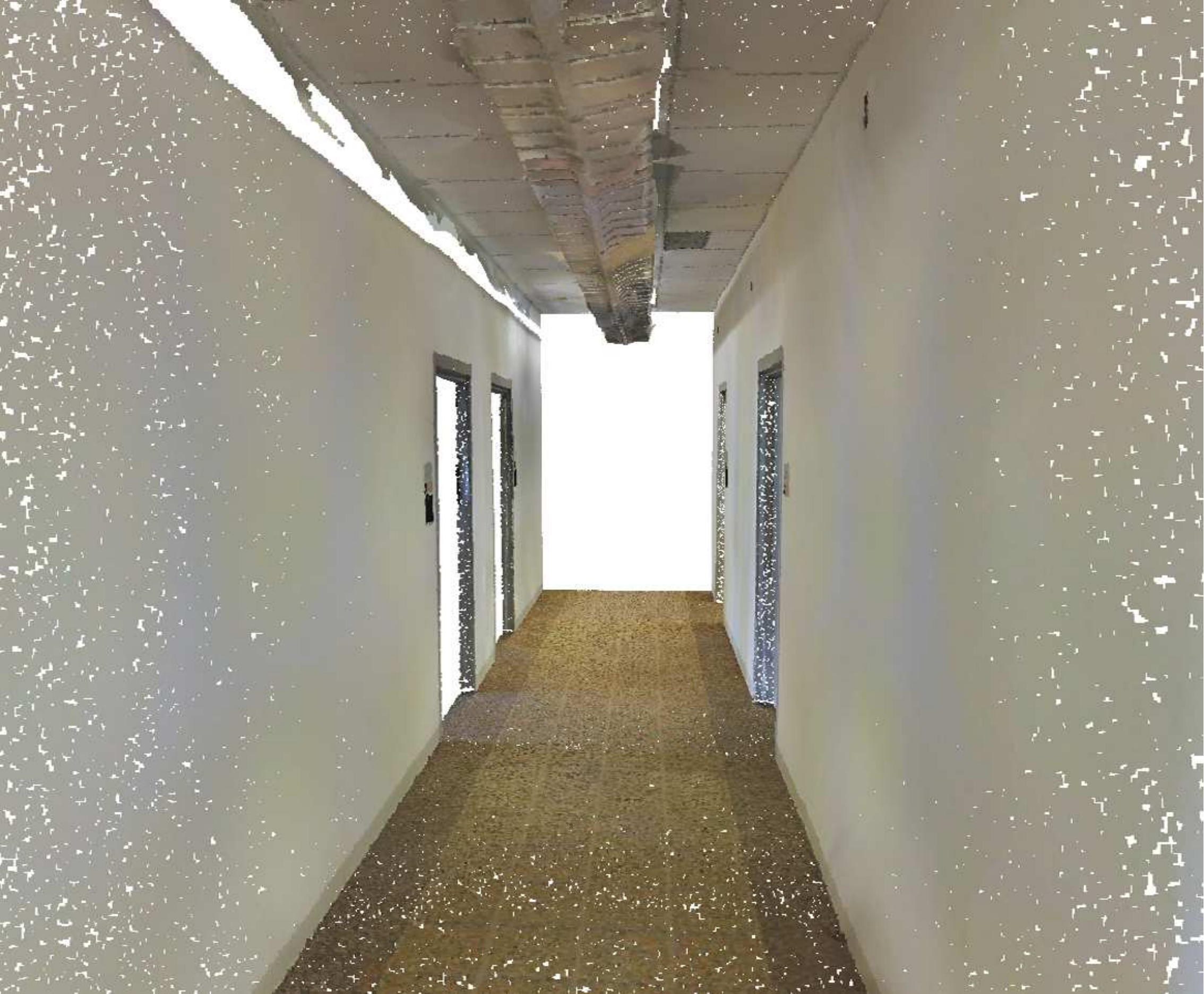}%
	\hspace{2pt}\includegraphics[width=\unit]{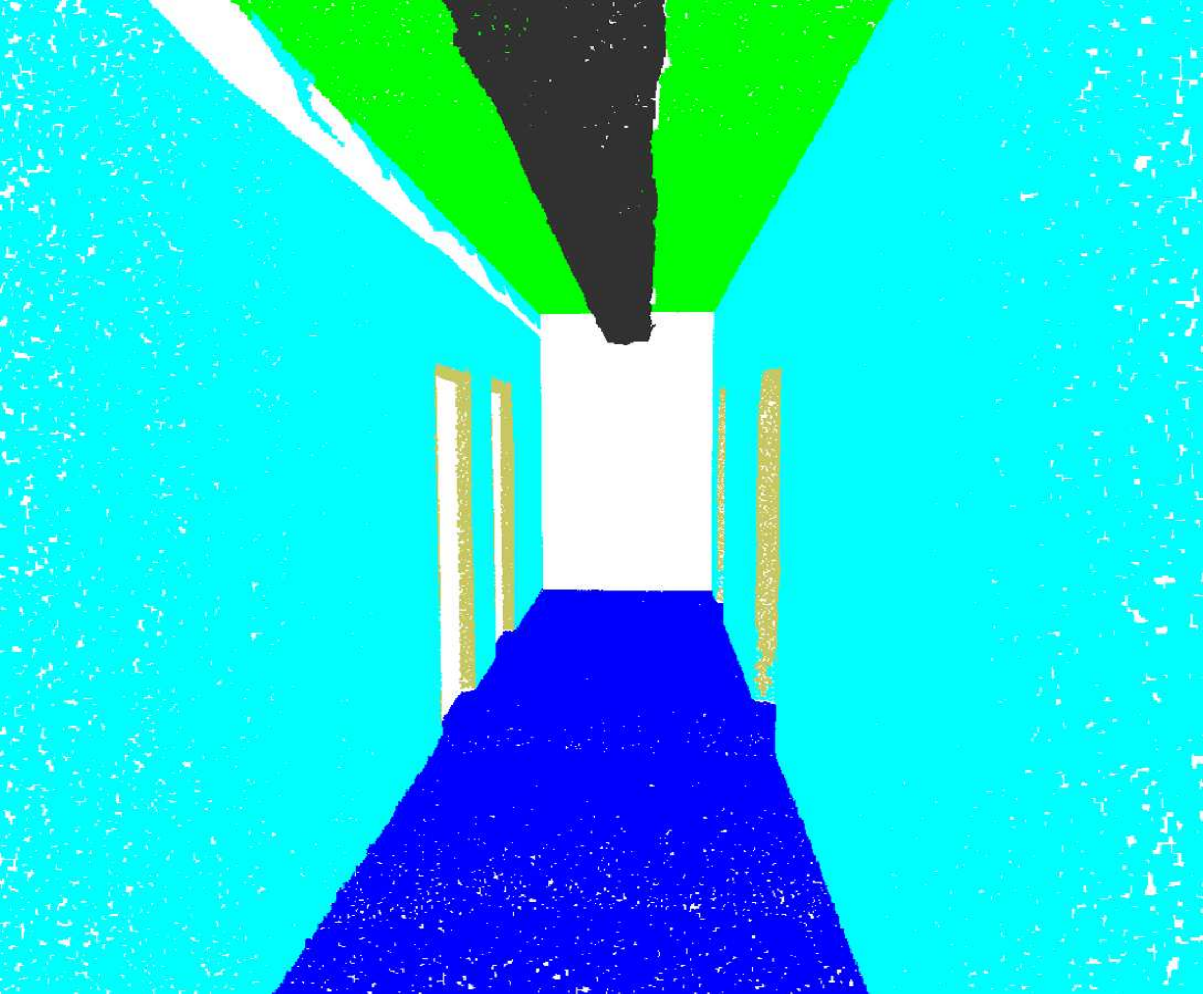}%
	\hspace{2pt}\includegraphics[width=\unit]{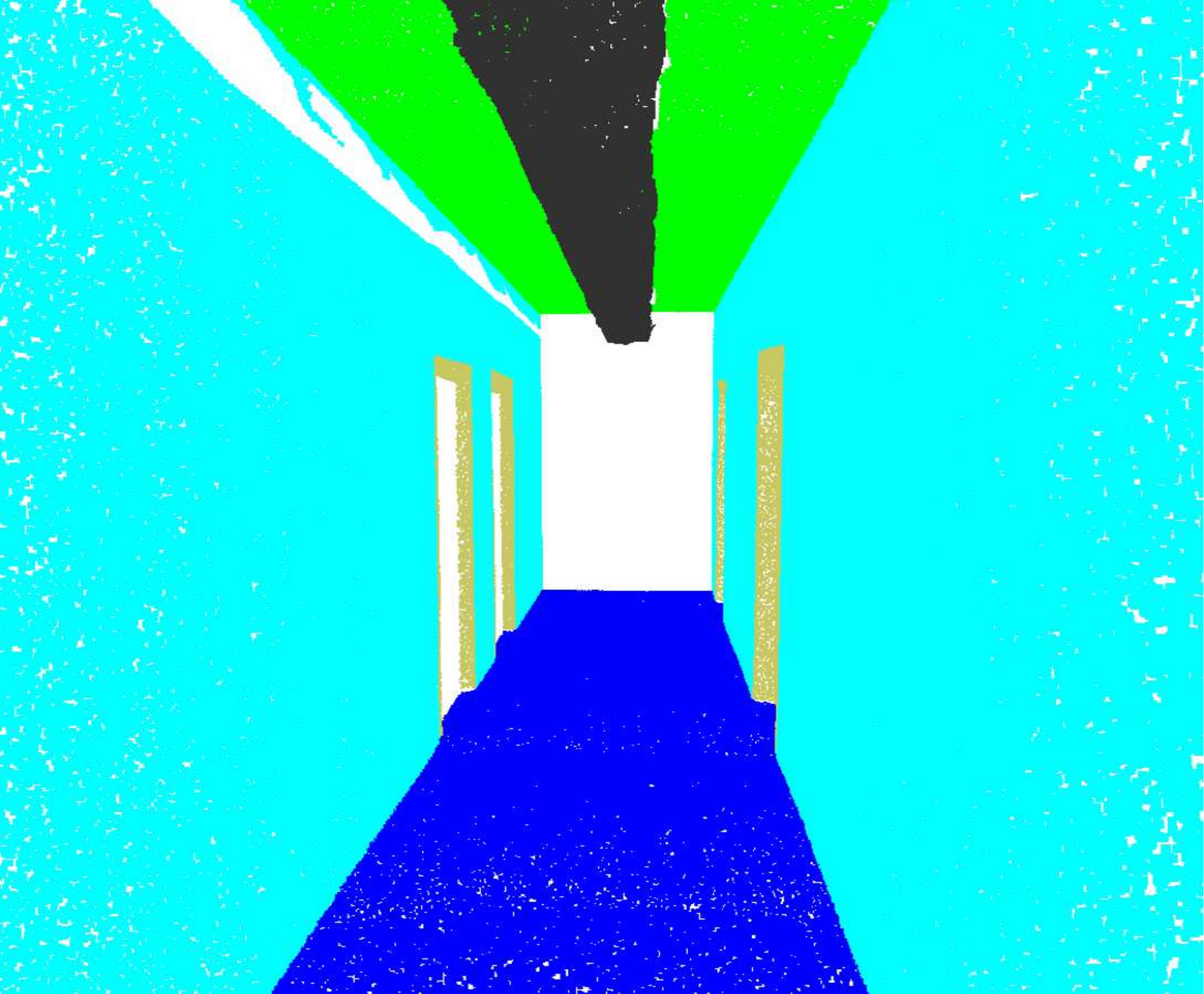}
	
	\vspace{-2pt}
	
	\subfigure[Input]{\includegraphics[width=\unit]{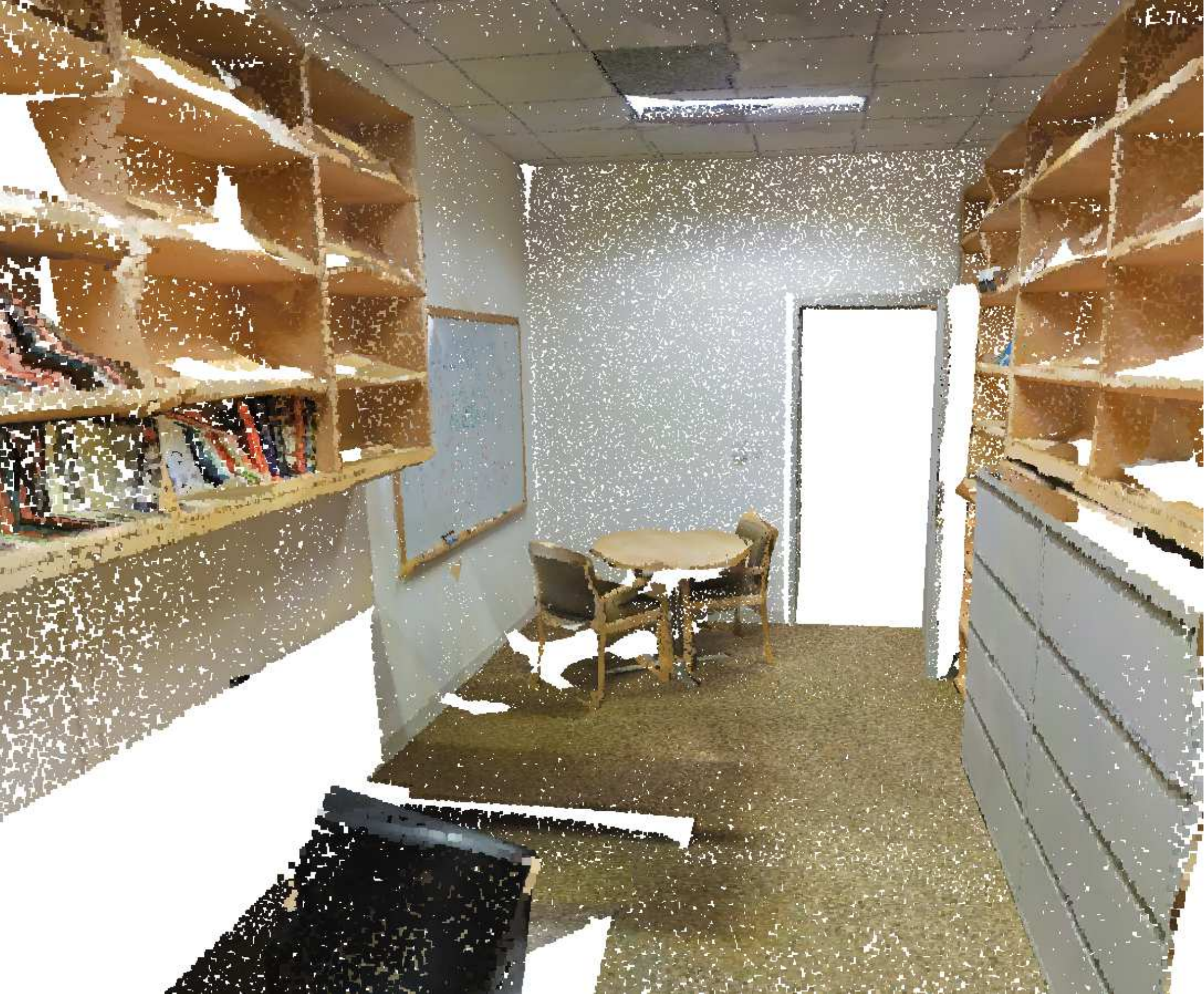}}%
	\hspace{2pt}\subfigure[Prediction]{\includegraphics[width=\unit]{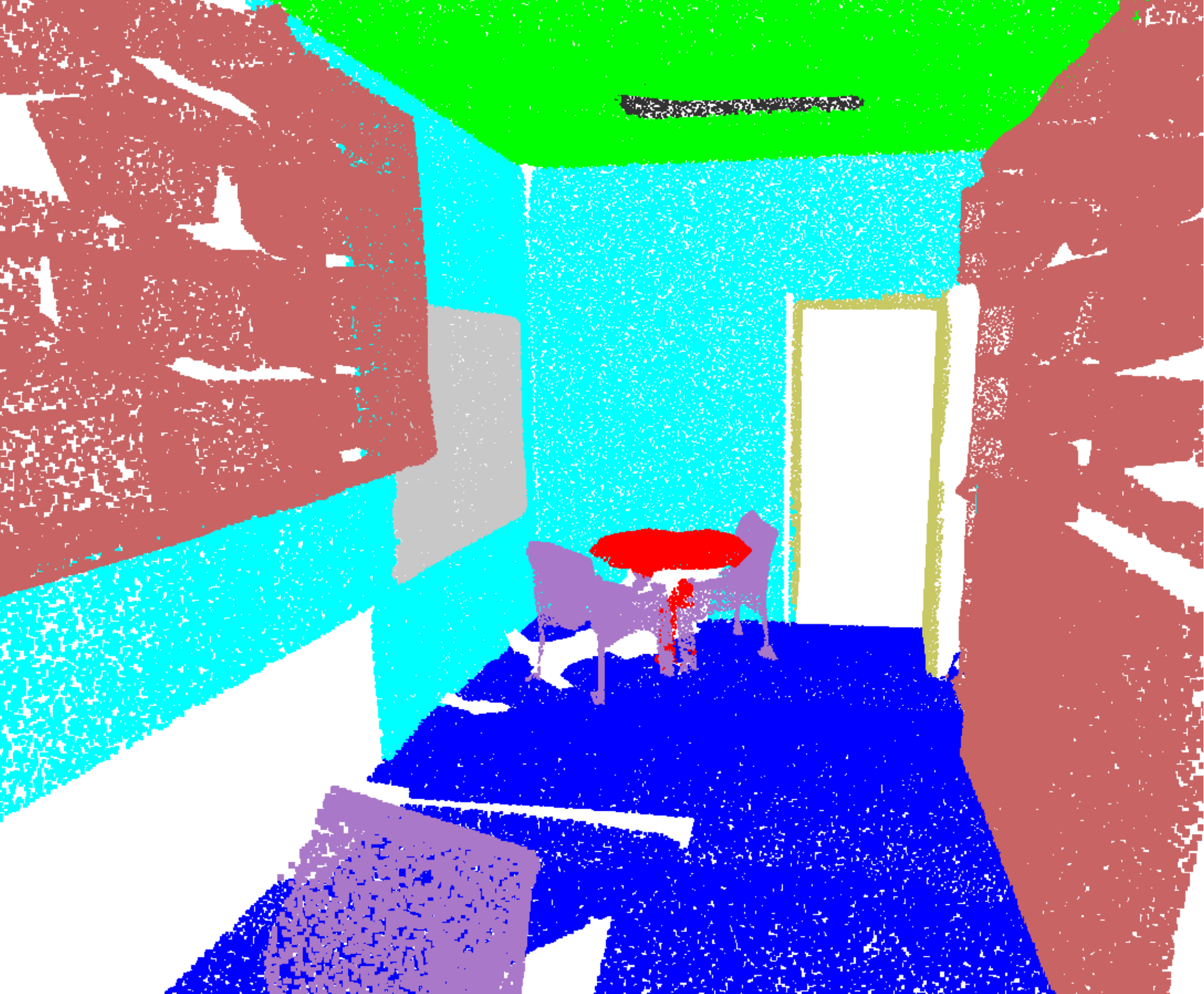}}%
	\hspace{2pt}\subfigure[Ground Truth]{\includegraphics[width=\unit]{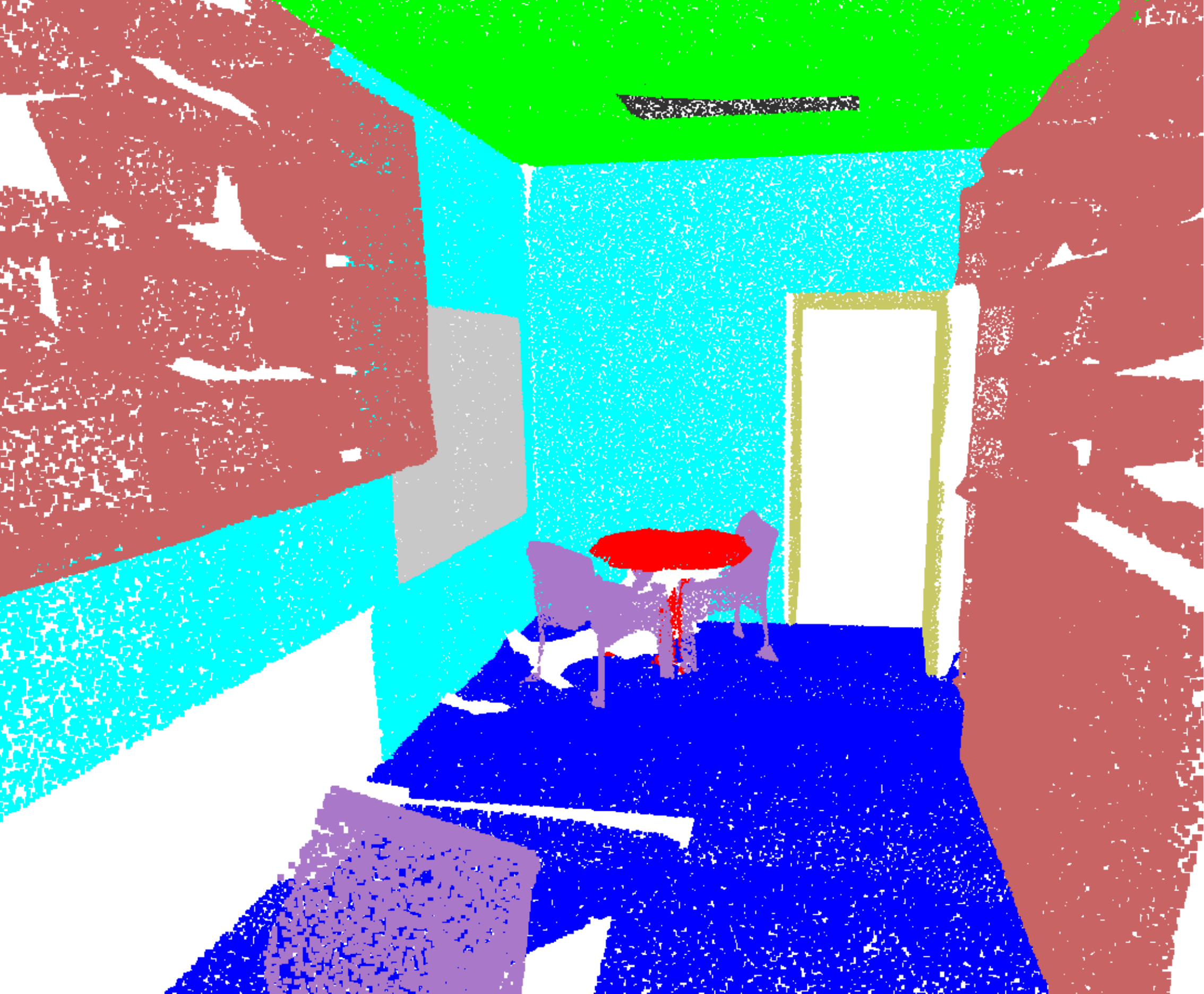}}
	
	\caption{Visualization of semantic segmentation results on the S3DIS dataset. We show the input point cloud, and labelled points mapped to RGB colors.}
	\label{fig:indoor}
\end{figure}

\textbf{Results.} We report the mean classwise intersection over union (mIoU), mean classwise accuracy (mAcc) and overall accuracy (OA) in Tab.~\ref{table:scene_results}. The IoU of each class is also provided. The proposed AdaptConv outperforms the state-of-the-arts in most of the categories, which further demonstrates the effectiveness of adaptive convolutions over fixed kernels. The qualitative results are visualized in Fig.~\ref{fig:indoor} where we show rooms from different areas of the building. Our method can correctly detect less obvious edges of, \eg, pictures and boards on the wall.

\subsection{Ablation studies}
\label{sec:eval:ablation}
In this section, we explain some of the architecture choices used in our network, and demonstrate the effectiveness of AdaptConv compared to several ablation networks.

\textbf{Adaptive convolution vs Fixed kernels.} We compare our AdaptConv with fixed kernel convolutions, including methods using the attention mechanism and standard graph convolution (DGCNN\cite{wang2019graph}), as discussed in the introduction. We train these models on ShapeNetPart dataset for segmentation, and design several ablation networks by replacing AdaptConv layers with fixed kernel layers and keeping other architectures the same.
%In order to tackle the isotropy of standard graph convolution based on a fixed kernel, several prior works \cite{velivckovic2017graph,wang2019graph,verma2018feastnet} utilize the attention mechanism. In this section, we compare our model with these graph attention networks (Attention). Since they may not focus on tasks of point cloud analysis, we design several ablations using attentional graph convolution layers in our framework to demonstrate the effectiveness of our adaptive convolution. 

Specifically, \cite{velivckovic2017graph} assign attentional weights to different neighboring points and \cite{wang2019graph} further designs a channel-wise attentional function. We use their layers and denote these two ablations as Attention Point and Attention Channel in Tab.~\ref{table:ab:seg} respectively. We only replace the AdaptConv layers in our network and the feature inputs $\Delta f_{ij}$ are the same as our model. Besides, we also show the result by using standard graph convolutions (GraphConv), which can be seen as a similar version of DGCNN \cite{wang2019dynamic}. From the comparison, we see that our method achieves better results than the fixed kernel graph convolutions.

\textbf{Feature decisions.} In AdaptConv, the adaptive kernel is generated from the feature input $\Delta f_{ij}$, and subsequently convolved with the corresponding  $\Delta x_{ij}$. %In this design, it serves as a function to establish the spatial relations between a pair of points in the neighborhood. 
Note that, in our experiments, $\Delta x_{ij}$ corresponds to the $(\mathbf{x}, \mathbf{y}, \mathbf{z})$ spatial coordinates of the points. We have discussed several other choices of $\Delta x_{ij}$ in Eq.~\ref{equ:convolution} in Sec.~\ref{sec:method:feature}, which can be evaluated by designing these ablations:

$\bullet$ Feature - In Eq.~\ref{equ:convolution}, we convolve the adaptive kernel $\hat{e}_{ijm}$ with their current point features. That is, $\Delta x_{ij}$ is replaced with $\Delta f_{ij}$ and the kernel function is $g_m: \mathbb{R}^{2D} \rightarrow \mathbb{R}^{2D}$. This makes the kernel learn to adapt to the features from previous layer and extracts the feature relations.

$\bullet$ Initial attributes - The point normals $(n_x, n_y, n_z)$ are included in the part segmentation task on ShapeNetPart, leading to a 6-dimensional initial feature attributes for each point. Thus, we design three ablations where we use only spatial inputs (Ours), only normal inputs (Normal) and both of them (Initial attributes). The kernel function is modified correspondingly.

The resulting IoU scores are shown in Tab.~\ref{table:ab:seg}. As one can see, $(\mathbf{x}, \mathbf{y}, \mathbf{z})$ is the most critical initial attribute (probably the only attribute) in point clouds, thus it is recommended to use them in the convolution with adaptive kernels. Althrough achieving a promising result, the computational cost for the Feature ablation is extremely high since the network expands heavily when it is convolved with a high-dimensional feature.

\begin{figure}[t]
	\centering
	
	\includegraphics[width=0.98\linewidth]{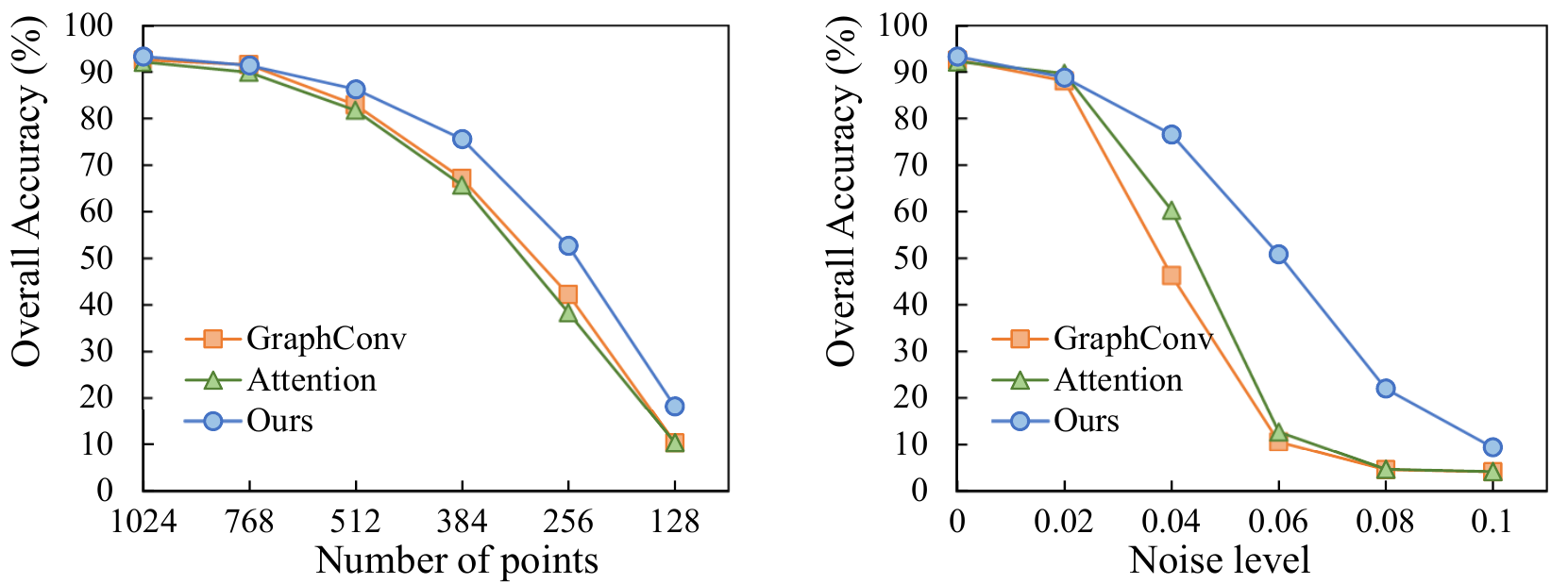}
	
	\caption{Robustness test on ModelNet40 for classification. GraphConv indicates the standard graph convolution network. Attention indicates the ablation where we replace the AdaptConv layers with graph attention layers (point-wise). From the comparison, we can see that our model is more robust to point density and noise perturbation.}
	\label{fig:robust}
\end{figure}

\begin{table}[t]
	\centering
	\small
	%\footnotesize
	\setlength{\tabcolsep}{5mm}
	\begin{tabular}{c|cc} %表格6列 全部居中显示
		\toprule[1pt]
		Number $k$ & mAcc(\%) & OA(\%) \\
		\midrule[0.3pt]
		\midrule[0.3pt]
		5						& 89.4 & 92.8 \\
		10						& 90.7 & 93.2 \\
		20						& 90.7 & 93.4 \\
		40						& 90.4 & 93.0 \\
		\bottomrule[1pt]
	\end{tabular}
	\vspace{5pt}
	\caption{Results of our classification network with different numbers $k$ of nearest neighbors.}
	\label{table:numberk}
\end{table}

\begin{figure*}
	\centering
	\newlength{\unitx}
	\setlength{\unitx}{0.14\linewidth}
	
	\includegraphics[width=\unitx]{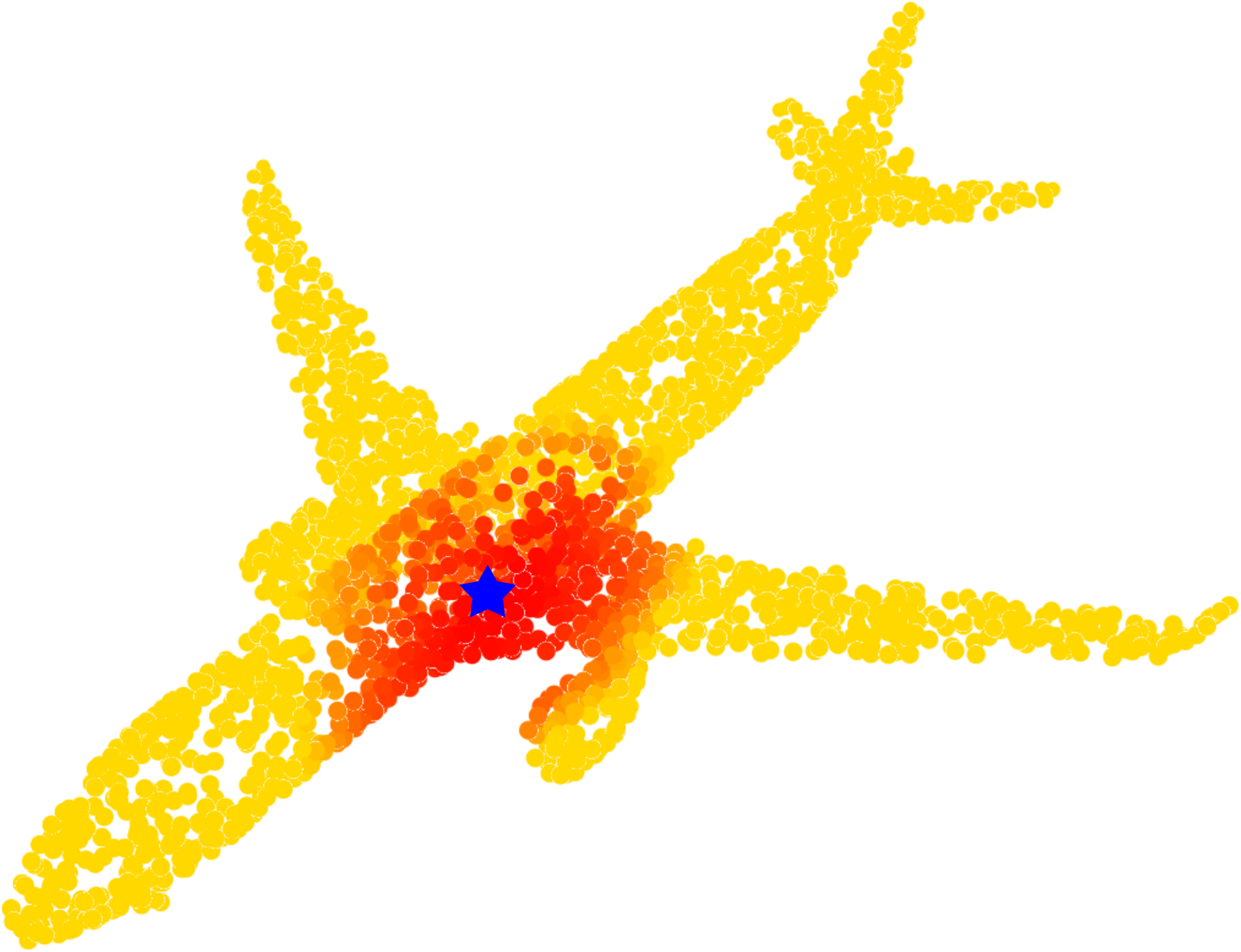}%
	\includegraphics[width=\unitx]{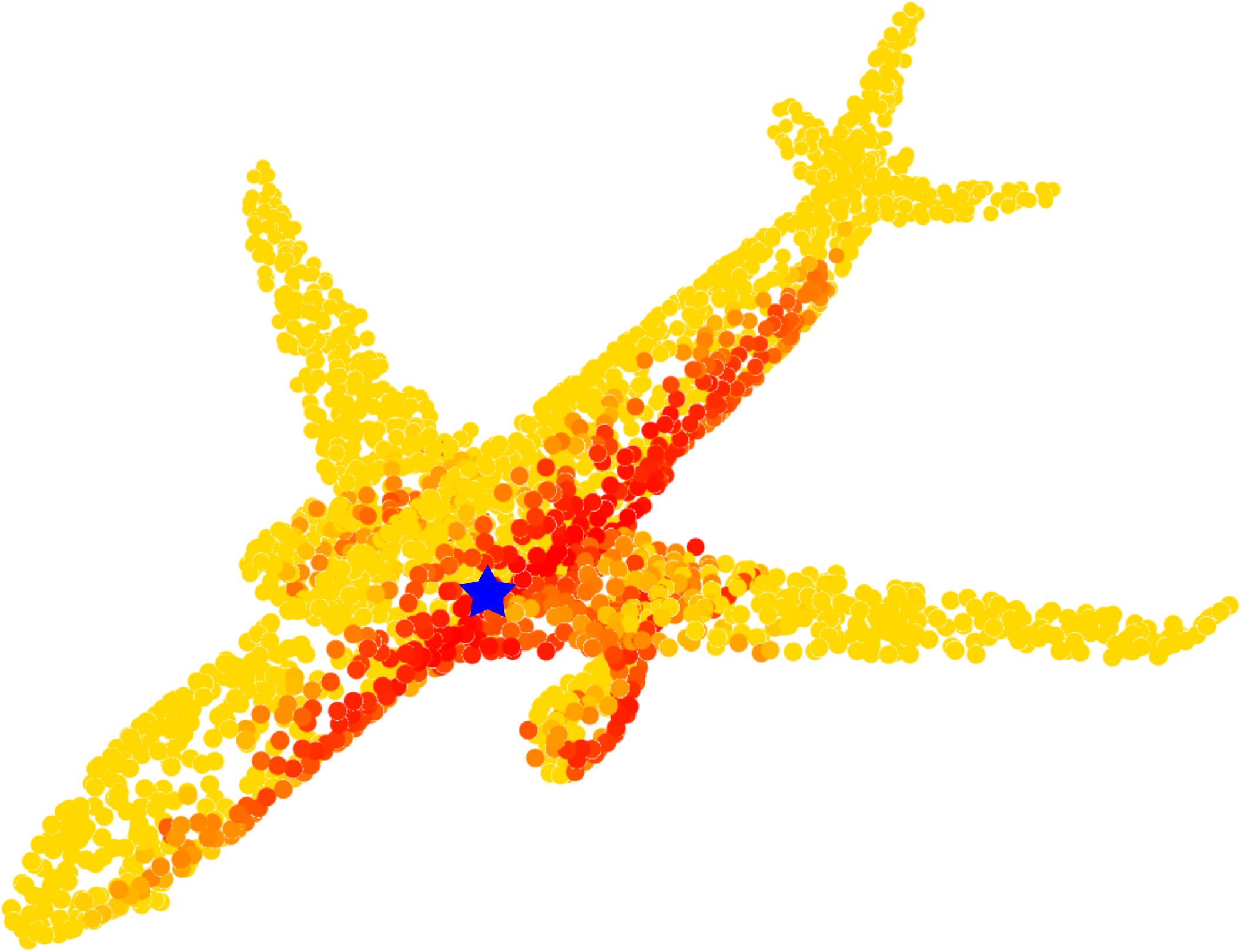}%
	\includegraphics[width=\unitx]{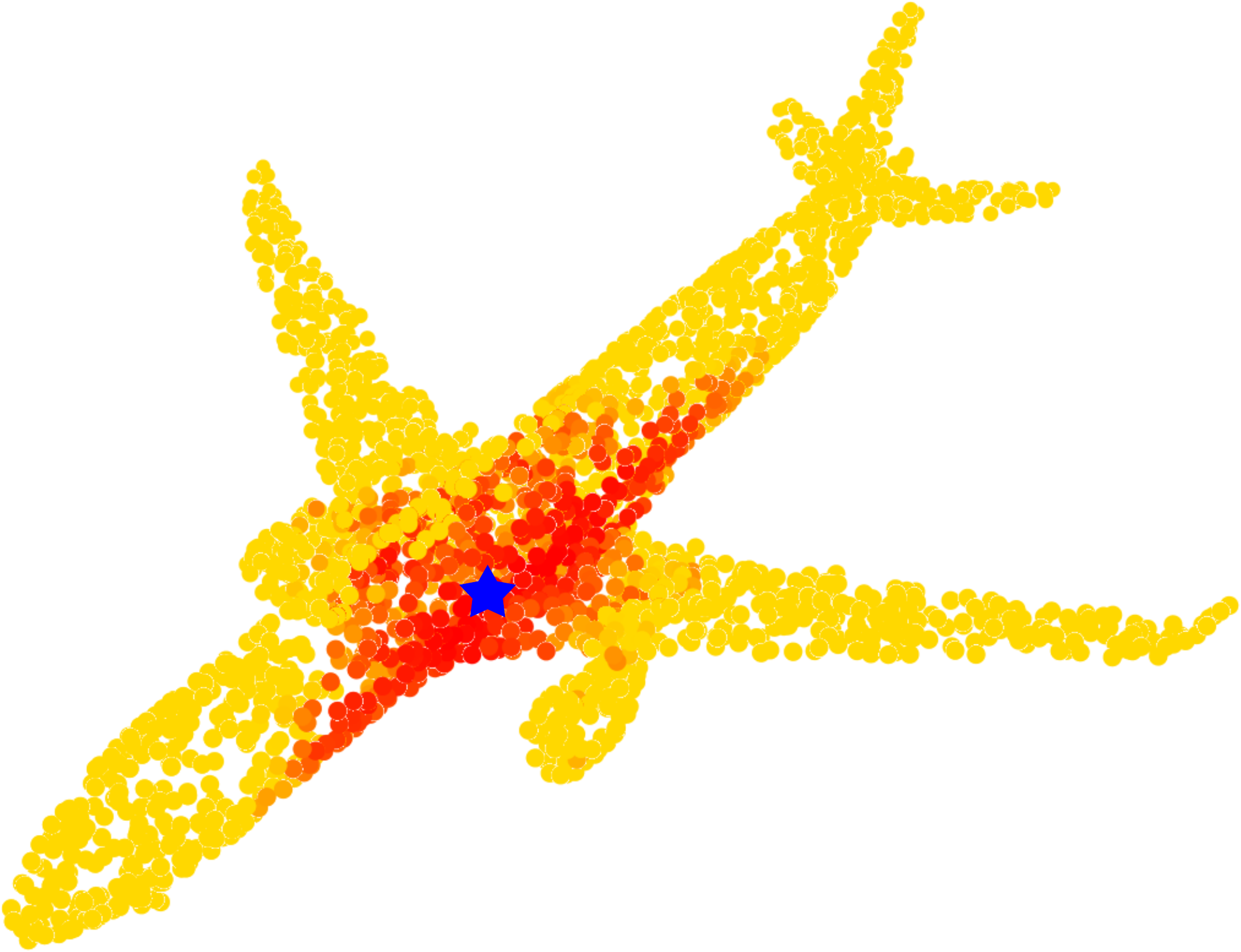}%
	\includegraphics[width=\unitx]{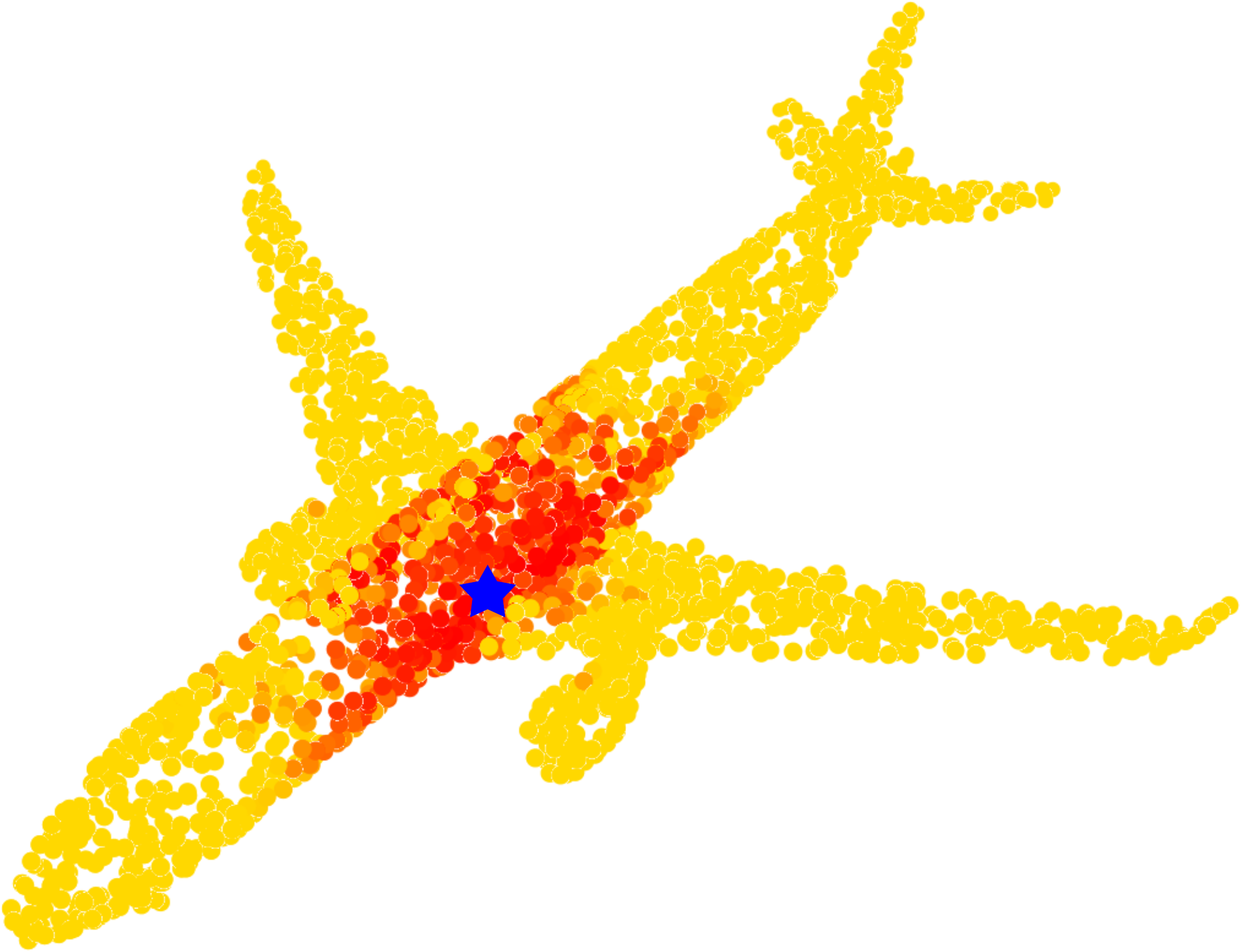}%
	\includegraphics[width=\unitx]{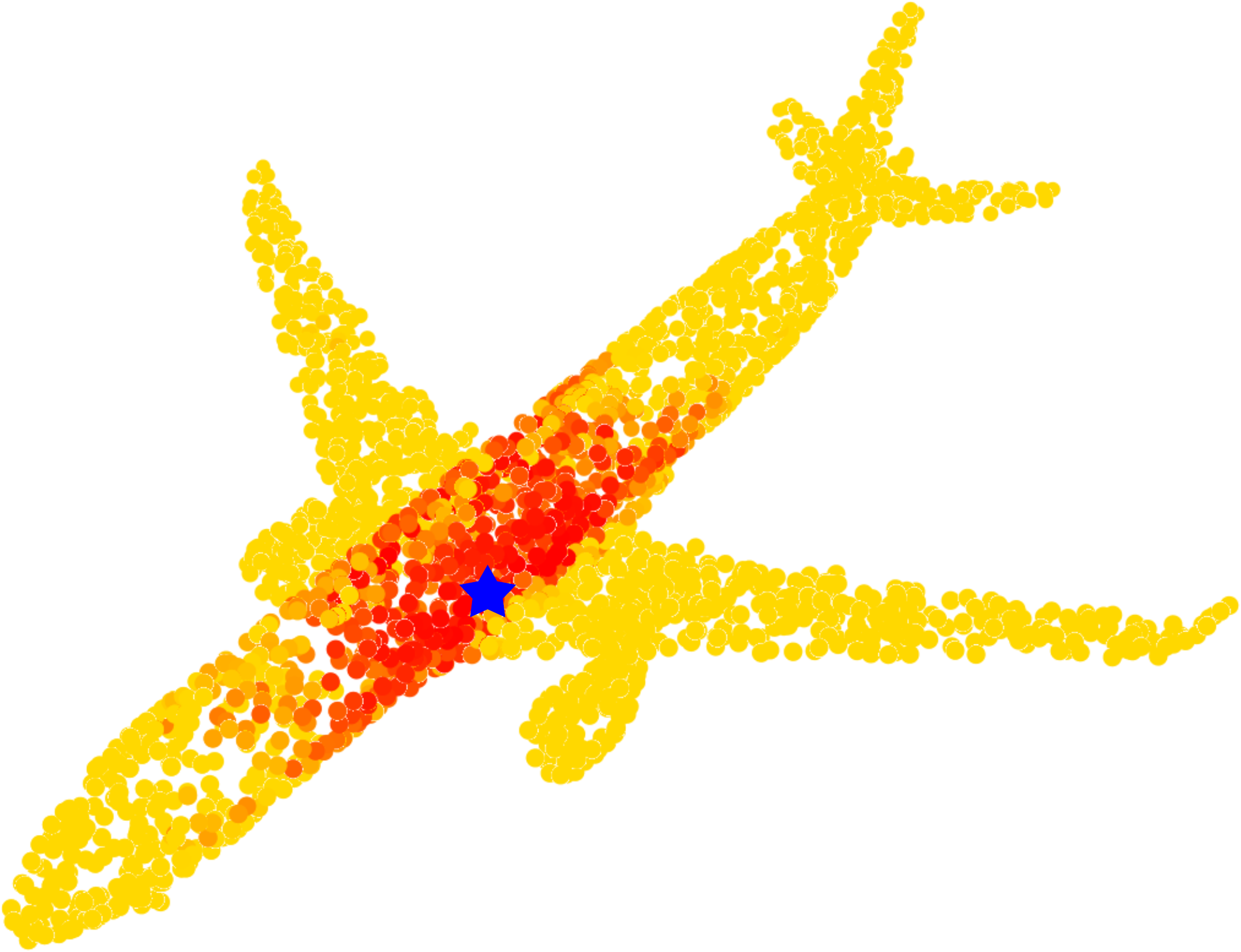}%
	\includegraphics[width=\unitx]{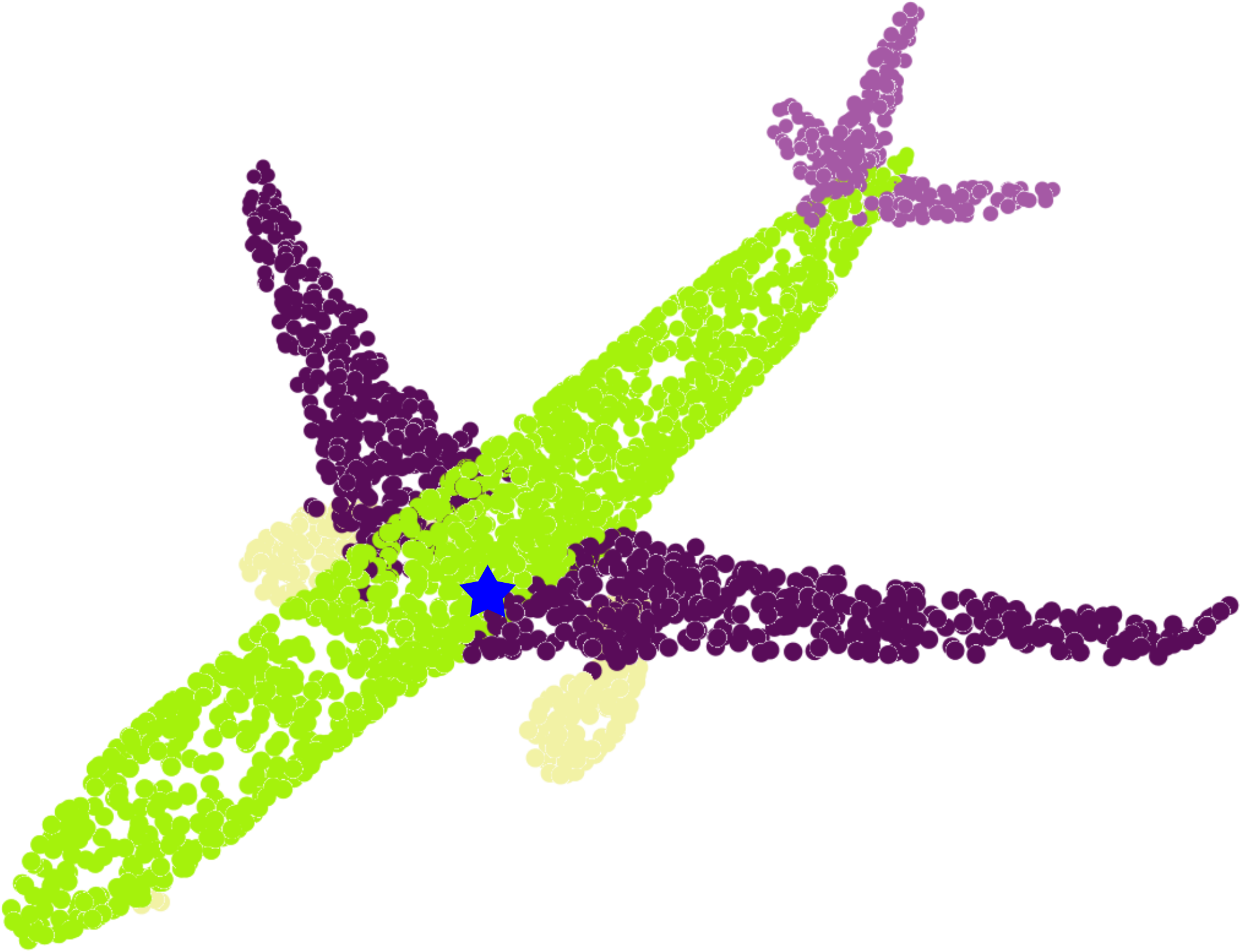}%
	\includegraphics[width=\unitx]{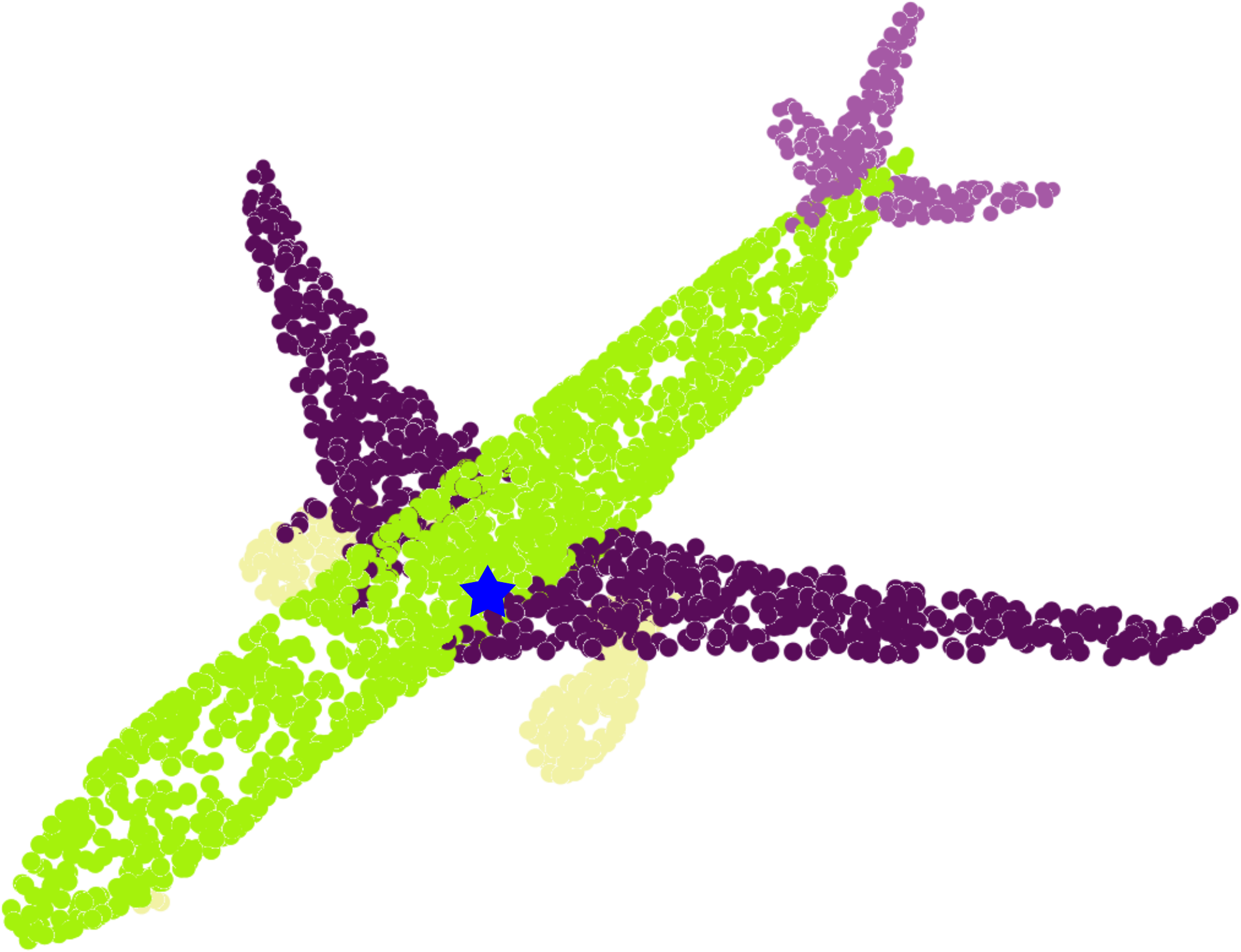}
	
	\subfigure[Spatial]{\includegraphics[width=\unitx]{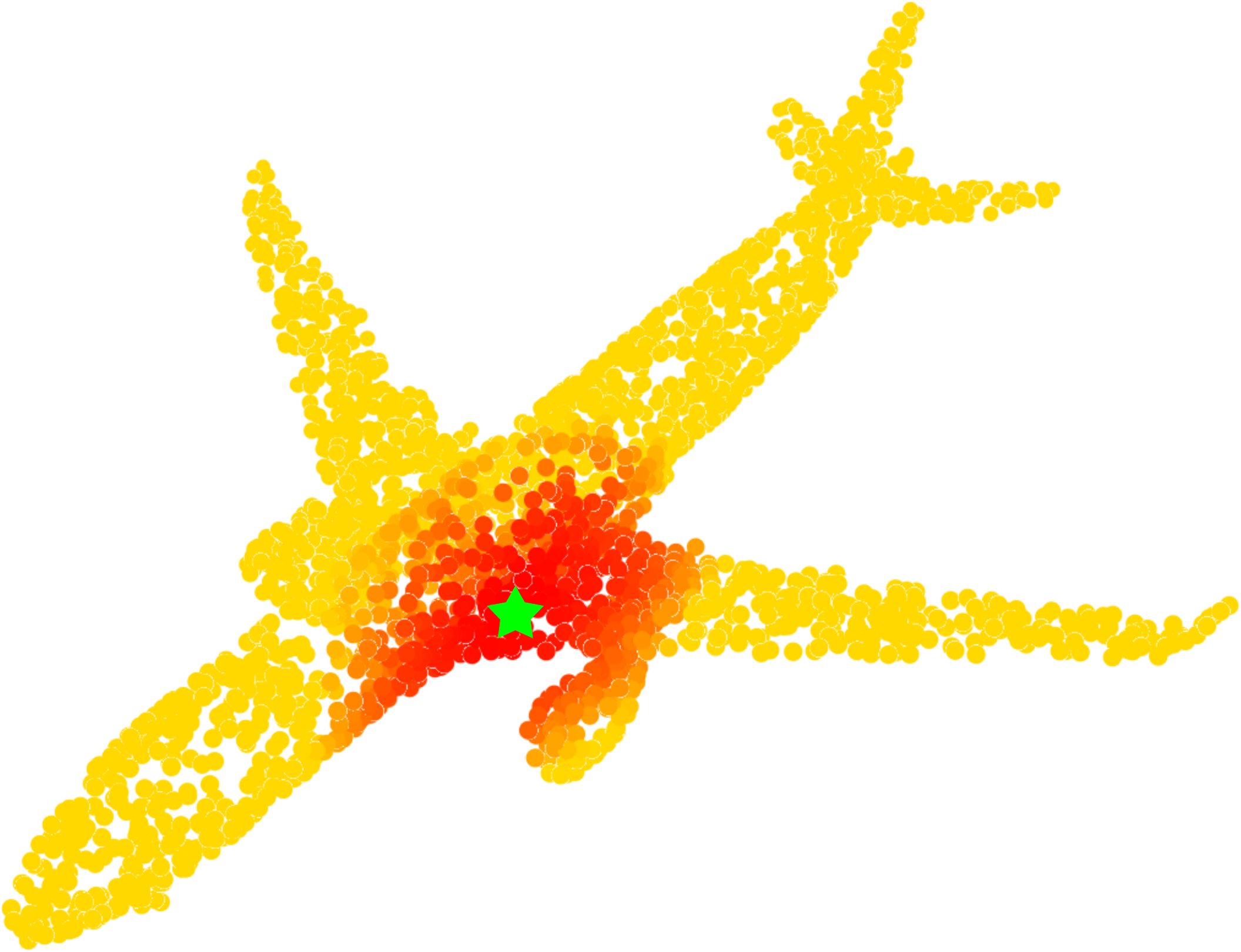}}%
	\subfigure[Layer1]{\includegraphics[width=\unitx]{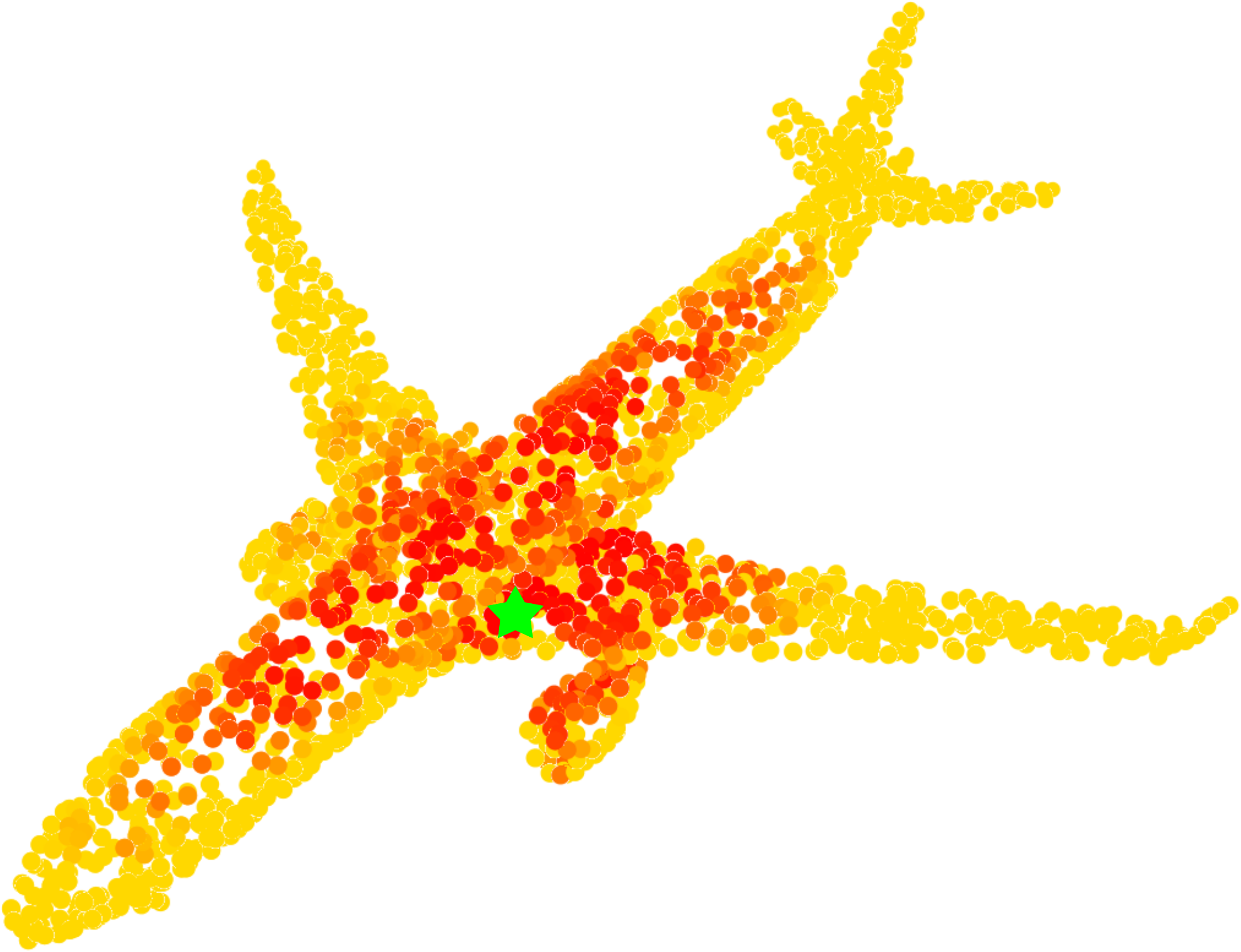}}%
	\subfigure[Layer2]{\includegraphics[width=\unitx]{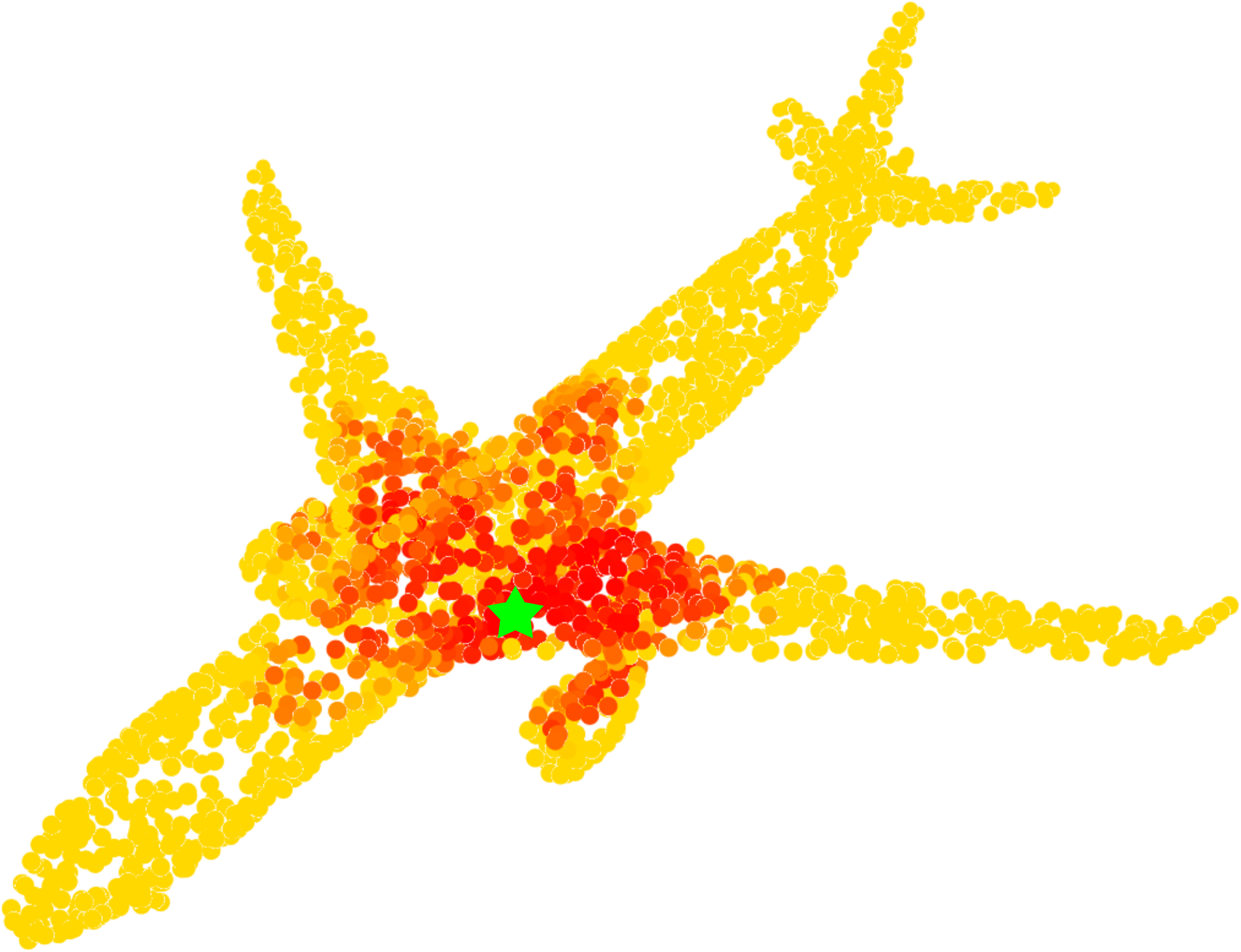}}%
	\subfigure[Layer3]{\includegraphics[width=\unitx]{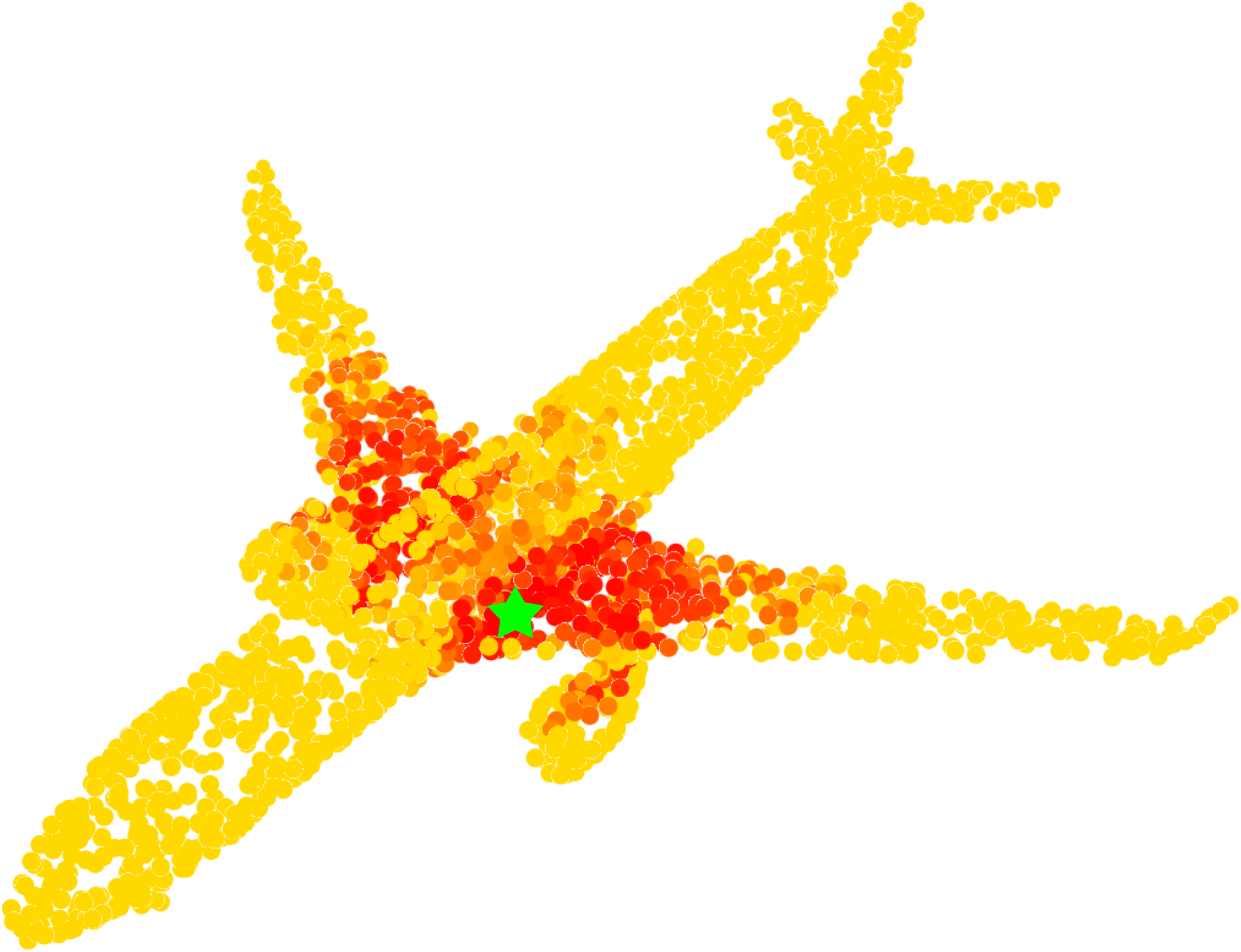}}%
	\subfigure[Layer4]{\includegraphics[width=\unitx]{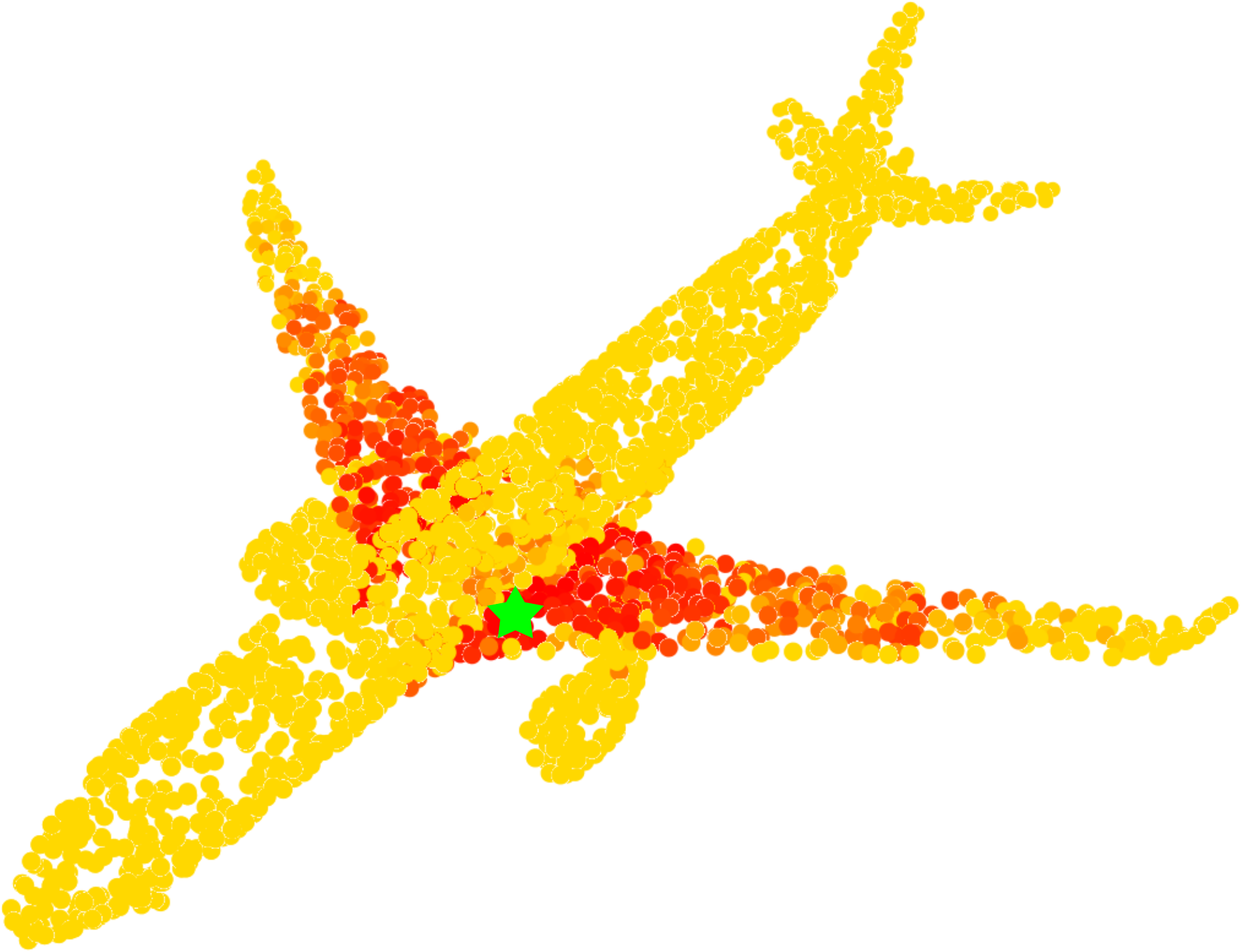}}%
	\subfigure[Ours]{\includegraphics[width=\unitx]{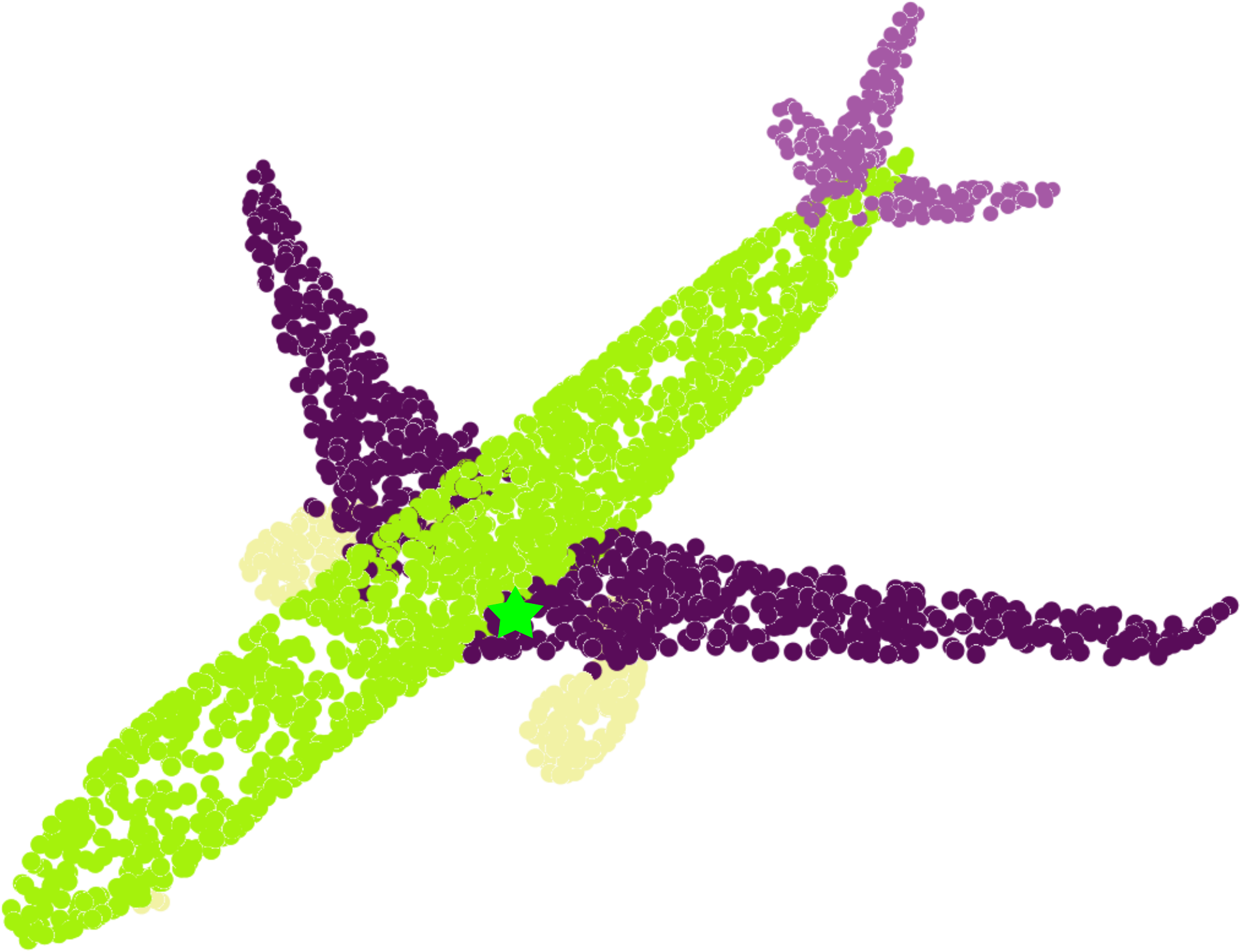}}%
	\subfigure[GT]{\includegraphics[width=\unitx]{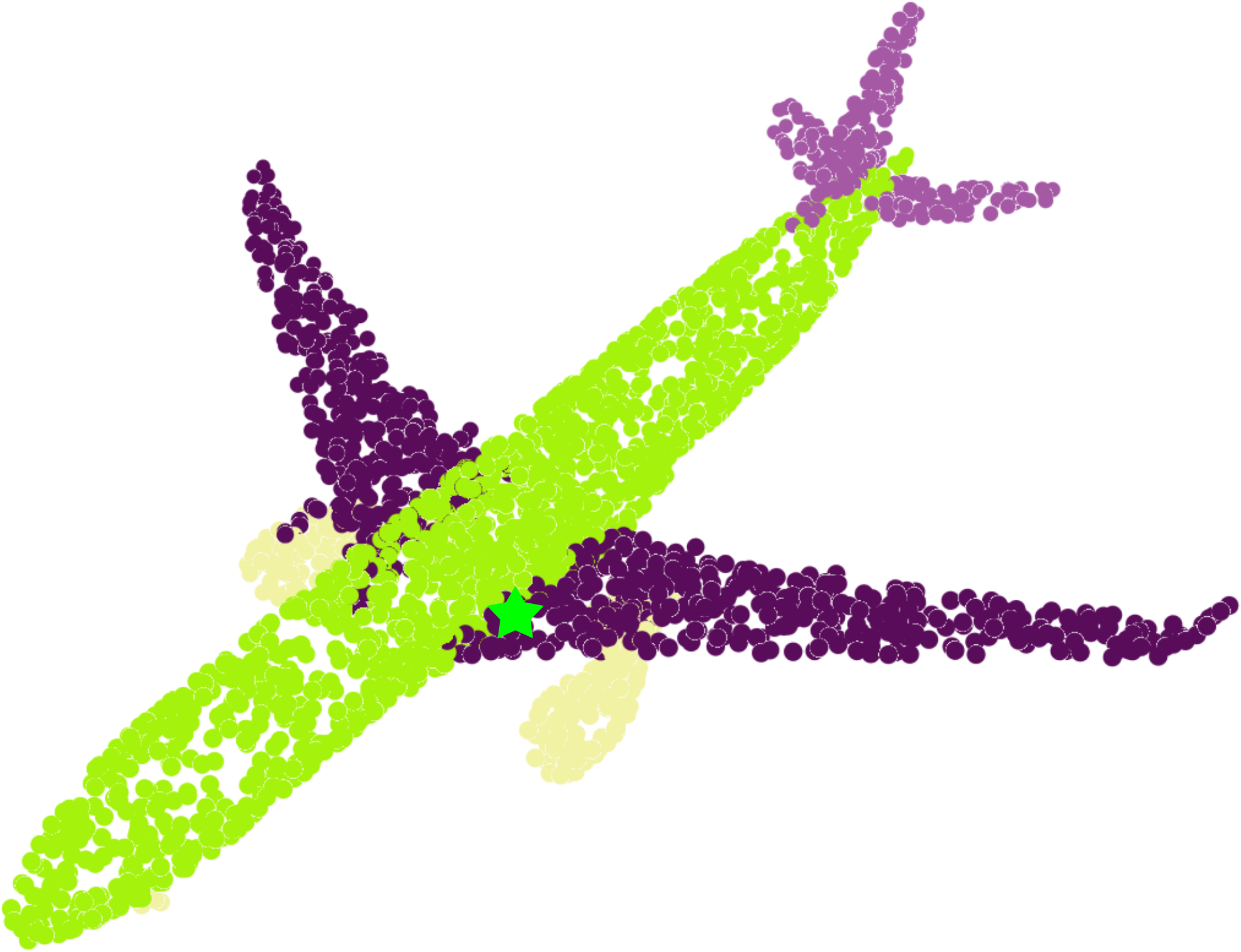}}
	
	\caption{Visualize the euclidean distances between two target points (blue and green stars) and other points in the feature space (red: near, yellow: far).  }
	\label{fig:insight1}
\end{figure*}

\subsection{Robustness test}
We further evaluate the robustness of our model to point cloud density and noise perturbation on ModelNet40 \cite{wu20153d}. We compare our AdaptConv with several other graph convolutions as discussed in Sec.~\ref{sec:eval:ablation}. All the networks are trained with 1k points and neighborhood size is set to $k=20$. In order to test the influence of point cloud density, a series of numbers of points are randomly dropped out during testing. For noise test, we introduce additional Gaussian noise with standard deviations according to the point cloud radius. From Fig.~\ref{fig:robust}, we can see that our method is robust to missing data and noise, thanks to the adaptive kernel in which the structural connections can be extracted dynamically in a sparser area.

Also, we experiment the influence of different numbers $k$ of the nearest neighboring points in Tab.~\ref{table:numberk}. We choose several typical sizes for testing. Reducing the number of neighboring points leads to less computational cost while the performance will degenerate due to the limitation of receptive field. Our network still achieves a promising result when $k$ is reduced to 5. On the other hand, with certain point density, a larger $k$ doesn't improve the performance since the local information dilutes within a larger neighborhood.

\begin{table}[t]
	\centering
	\small
	%\footnotesize
	\setlength{\tabcolsep}{3.5mm}
	\begin{tabular}{c|cc} %表格6列 全部居中显示
		\toprule[1pt]
		Method & \#parameters & OA(\%) \\
		\midrule[0.3pt]
		\midrule[0.3pt]
		PointNet \cite{QiSMG17}							& 3.5M & 89.2 \\
		PointNet++ \cite{qi2017pointnet++}				& 1.48M & 91.9 \\
		DGCNN \cite{wang2019dynamic}					& 1.81M & 92.9 \\
		KPConv \cite{thomas2019kpconv}					& 14.3M & 92.9 \\
		Ours											& 1.85M & \textbf{93.4} \\
		\bottomrule[1pt]
	\end{tabular}
	\vspace{5pt}
	\caption{The number of parameters and overall accuracy of different models.}
	\label{table:complexity}
\end{table}

\subsection{Efficiency}
\label{sec:eval:eff}
To compare the complexity of our model with previous state-of-the-arts, we show the parameter numbers and the corresponding results of networks in Tab.~\ref{table:complexity}. These models are based on ModelNet40 for classification task. From the table, we see that our model achieves the best performance of 93.4\% overall accuracy and the model size is relatively small. Compared with DGCNN \cite{wang2019dynamic} which can be seen as a standard graph convolution version in our ablation studies, the proposed adaptive kernel performs better while being efficient.

%\textbf{Effectiveness of AdaptConv.} The proposed AdaptConv can be easily integrated into existing graph-based networks to improve their performance. 

\section{Visualization and learned features}
%To further demonstrate the advantages of our method, we visualize the segmentation results on ShapeNetPart dataset in Fig.~\ref{fig:partseg}. In this experiment, we compare the results from DGCNN \cite{wang2019dynamic}, the Attention ablation (Point) in Sec.~\ref{sec:eval:ablation} and our model. The AdaptConv segmentation results are better in challenging regions, such as part boundaries and object edges. This verifies that our method is able to capture distinguishable features for points belonging to different parts.
%In Fig.~\ref{fig:insight1}, we show more results of point euclidean distances in the feature space. Two target points are selected which belong to different parts of the object. While being spatially close, our network can capture their different geometric characteristics and find the sematically similar points.

To achieve a deeper understanding of AdaptConv, we explore the feature relations in several intermediate layers of the network to see how AdaptConv can distinguish points with similar spatial inputs. In this experiment, we train our model on ShapeNetPart dataset for segmentation. In Fig.~\ref{fig:insight1}, two target points (blue and green stars in 1-st and 2-nd rows respectively) are selected which belong to different parts of the object. We then compute the euclidean distances to other points in the feature space, and visualize them by coloring the points with similar learned features in red. We can see that, while being spatially close, our network can capture their different geometric characteristics and segment them properly. Also, from the 2-nd row of Fig.~\ref{fig:insight1}, points belonging to the same semantic part (the wings) share similar features while they may not be spatially close. This shows that our model can extract valuable information in a non-local manner.

%As shown in Fig.~\ref{fig:insight}, a target point (green point) is picked up and we compute the distances in the feature space to other points. When the point is close to boundaries and edges, our network encourages the points in its neighborhood to have distinguishable features. Thus, it is separated from other parts of the objects, as shown in the first row of Fig.~\ref{fig:insight}.  Also, we see that in the second row of Fig.~\ref{fig:insight}, points belonging to the same semantic part share similar features while they may not be spatially close. Therefore, the model can extract semantically valuable information in a non-local manner, thanks to the dynamic graph updating scheme. Note that, Fig.~\ref{fig:insight:spatial} indicates the spatial distances with regard to the central point.
\label{sec:eval}

\section{Conclusion}
In this paper, we propose a novel adaptive graph convolution (AdaptConv) for 3D point cloud. The main contribution of our method lies in the designed adaptive kernel in the convolution, which is dynamically generated according to the point features. Instead of using a fixed kernel that captures correspondences indistinguishably between points, our AdaptConv can produce learned features that are more flexible to shape geometric structures. We have applied AdaptConv to train end-to-end deep networks for several point cloud analysis tasks, outperforming the state-of-the-arts on several public datasets. Further, AdaptConv can be easily integrated into existing graph CNNs to improve their performance by simply replacing the existing kernels with the adaptive kernels.
\label{sec:conclusion}

\vspace{5pt}
\noindent\textbf{Acknowledgements.}
This work was supported by the National Natural Science Foundation of China (No. 62032011, No.62172218, No. 61672273) and the Research Grants Council of the Hong Kong Special Administrative Region, China (Project No. PolyU 152035/17E and 15205919).

%This work was supported by the National Natural Science Foundation of China (No. 61502137), the Hong Kong Research Grants Council (No. PolyU 152035/17E), the HKIBS Research Seed Fund 2019/20 (No. 190-009), and the Research Seed Fund (No. 102367) of Lingnan University, Hong Kong.

{\small
\bibliographystyle{ieee_fullname}
\bibliography{egbib}
}

\end{document}